\documentclass[10pt,twocolumn,letterpaper]{article}

\usepackage{cvpr}
\usepackage{times}
\usepackage{epsfig}
\usepackage{graphicx}
\usepackage{amsmath}
\usepackage{amssymb}

\usepackage{booktabs}       %
\usepackage{amsfonts}       %
\usepackage{nicefrac}       %
\usepackage{color}
\usepackage{multirow}
\usepackage[super]{nth}
\usepackage{subfig}
\usepackage[normalem]{ulem} %

\newcommand{\myparagraph}[1]{\vspace{5pt}\noindent{\bf #1}}

\usepackage{multirow, makecell}

\usepackage[breaklinks=true,bookmarks=false]{hyperref}

\cvprfinalcopy %

\def\cvprPaperID{5612} %

\begin{document}

\title{Adversarial Inference for Multi-Sentence Video Description}

\author{Jae Sung Park$^{1}$, Marcus Rohrbach$^{2}$, Trevor Darrell$^{1}$, Anna Rohrbach$^{1}$ \\
$^1$ University of California, Berkeley, $^2$ Facebook AI Research}

\maketitle
\begin{abstract}
While significant progress has been made in the image captioning task, video description is still in its infancy due to the complex nature of video data. Generating multi-sentence descriptions for long videos is even more challenging. Among the main issues are the fluency and coherence of the generated descriptions, and their relevance to the video. Recently, reinforcement and adversarial learning based methods have been explored to improve the image captioning models; however, both types of methods suffer from a number of issues, \eg poor readability and high redundancy for RL and stability issues for GANs. In this work, we instead propose to apply adversarial techniques during inference, designing a discriminator which encourages better multi-sentence video description. In addition, we find that a multi-discriminator ``hybrid'' design, where each discriminator targets one aspect of a description, leads to the best results. Specifically, we decouple the discriminator to evaluate on three criteria: 1) visual relevance to the video, 2) language diversity and fluency, and 3) coherence across sentences. Our approach results in more accurate, diverse, and coherent multi-sentence video descriptions, as shown by automatic as well as human evaluation on the popular ActivityNet Captions dataset. %
\end{abstract}

\section{\label{sec:intro} Introduction}

\begin{figure}[t]
\scriptsize
\begin{center}
\includegraphics[width=\linewidth]{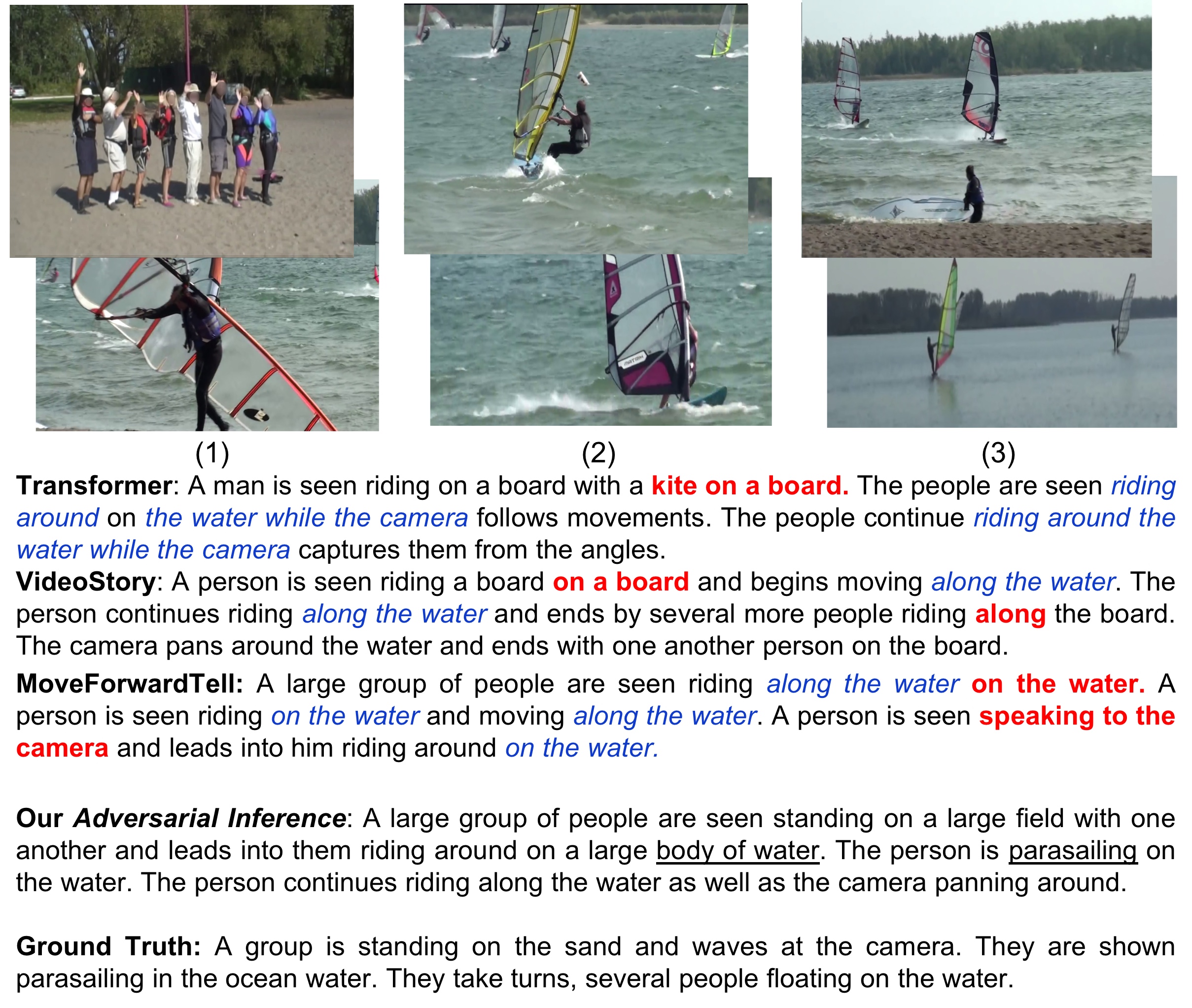} %
\vspace{-10pt}
\caption{Comparison of the state-of-the-art video description approaches, Transformer \cite{zhou2018end}, VideoStory \cite{gella2018dataset}, MoveForwardTell \cite{xiong2018move}, and our proposed \emph{Adversarial Inference}. Our approach generates more interesting and accurate descriptions with less redundancy. Video from ActivityNet Captions~\cite{caba2015activitynet,krishna2017dense} with three segments (left to right); red/bold indicates content errors, blue/italic indicates repetitive patterns, underscore highlights more interesting phrases.\vspace{-12pt}}
\label{fig:teaser}
\end{center}
\end{figure}

Being able to automatically generate a natural language description for a video has fascinated researchers since the early 2000s \cite{kojima02ijcv}. Despite the high interest in this task and ongoing emergence of new datasets \cite{gella2018dataset,krishna2017dense,zhou2018towards} and approaches \cite{xiong2018move,yu2018fine,zhou2018end}, it remains a highly challenging problem. Consider the outputs of the three recent video description methods on an example video from the ActivityNet Captions dataset~\cite{caba2015activitynet,krishna2017dense} in Figure \ref{fig:teaser}. We notice that there are multiple issues with these descriptions, in addition to the errors with respect to the video content: there are semantic inconsistencies and lack of diversity within sentences, as well as redundancies across sentences. There are multiple challenges towards more accurate and natural video description. One of the issues is the size of the available training data, which, despite the recent progress, is limited. Besides, video representations are more complex than \eg image representations, and require modeling temporal structure jointly with the semantics of the content. Moreover, describing videos with multiple sentences, requires correctly recognizing a sequence of events in a video, maintaining linguistic coherence and avoiding redundancy.
Another important factor is the target metric used in the description models. Most works still exclusively rely on the automatic metrics, e.g. METEOR \cite{meteor}, despite the evidence that they \emph{are not consistent} with human judgments \cite{kilickaya2016re,cider}. Further, some recent works propose to explicitly optimize for the sentence metrics using reinforcement learning based methods \cite{liu2017improved,rennie2017self}. These techniques have become quite widespread, both for image and video description \cite{anderson2018bottom,xiong2018move}. Despite getting higher scores, reinforcement learning based methods have been shown to lead to \emph{unwanted artifacts}, such as ungrammatical sentence endings \cite{guo2018improving}, increased object hallucination rates \cite{rohrbach2018emnlp} and lack of diverse content \cite{melnyk2018improved}. Overall, while informative, sentence metrics should not be the only way of evaluating the description approaches. 

Some works aim to overcome this issue by using the adversarial learning \cite{dai2017towards,shetty2017speaking}. While Generative Adversarial Networks \cite{goodfellow2014generative} have achieved impressive results for image and even video generation \cite{isola2017image,reed2016generative,wang2018vid2vid,CycleGAN2017}, their success in language generation has been limited \cite{subramanian2017adversarial,yu2017seqgan}. The main issue is the difficulty of achieving stable training due to the discrete output space \cite{caccia2018language,che2017maximum}. Another reported issue is lack of coherence, especially for long text generation \cite{holtzman2018learning}.
Still, the idea of \emph{learning} to distinguish the ``good'' natural descriptions from the ``bad'' fake ones, is very compelling.

Rather than learning with adversarial training, we propose a simpler approach, \emph{Adversarial Inference} for video description, which relies on a discriminator to improve the description quality. Specifically, we are interested in the task of multi-sentence video description \cite{rohrbach14gcpr,yu16cvpr}, \ie the output of our model is a paragraph that describes a video. We assume that the ground-truth temporal segments are given, \ie we do not address the event detection task, but focus on obtaining a coherent multi-sentence description. %
We first design a strong baseline generator model trained with the maximum likelihood objective, which relies on a previous sentence as context, similar to \cite{gella2018dataset,xiong2018move}. We also introduce object-level features in the form of object detections \cite{anderson2018bottom} to better represent people and objects in video. 
We then make the following contributions:

(1) We propose the \emph{Adversarial Inference} for video description, 
where we progressively sample sentence candidates for each clip, and select the best ones based on a discriminator's score. Prior work has explored sampling with log probabilities \cite{donahue17pami}, while we show that a specifically trained discriminator leads to better results in terms of correctness, coherence, and diversity (see Figure~\ref{fig:teaser}).

(2) Specifically, we propose the ``hybrid discriminator'', which combines three specialized discriminators: one measures the language characteristics of a sentence, the second assesses its relevance to a video segment, and the third measures its coherence with the previous sentence. Prior work has considered a ``single discriminator'' for adversarial training to capture both the linguistic characteristics and  visual relevance \cite{shetty2017speaking,dai2017towards}. We show that our ``hybrid discriminator'' outperforms the ``single discriminator'' design.

(3) We compare our proposed approach to multiple baselines on a number of metrics, including automatic sentence scores, diversity and repetition scores, person correctness scores, and, most importantly, human judgments. We show that our \emph{Adversarial Inference} approach leads to more accurate and diverse multi-sentence descriptions, outperforming GAN and RL based approaches in a human evaluation.

\section{\label{sec:related} Related Work}
We review  existing approaches to video description, including  recent work based on reinforcement and adversarial learning. We then discuss related works that also sample and re-score sentence descriptions, and some that aim to design alternatives to automatic evaluation metrics.

\myparagraph{Video description.}
Over the past years there has been an increased interest in video description generation, notably with the broader adoption of the deep learning techniques. S2VT \cite{venugopalan15iccv} was among the first approaches based on LSTMs \cite{hochreiter1997long,donahue15cvpr}; some of the later ones include \cite{pan16cvpr,rohrbach15gcpr,shetty16mm,yao2015iccv,youngjaeyu17cvpr,zanfir16accv}. Most recently, a number of approaches to video description have been proposed, such as replacing LSTM with a Transformer Network \cite{zhou2018end}, introducing a reconstruction objective \cite{wang2018reconstruction}, using bidirectional attention fusion for context modeling \cite{wang2018bidirectional}, and others \cite{chen2018less,gella2018dataset,li2018jointly}.

While most works focus on ``video in - one sentence out'' task, some aim to generate a multi-sentence paragraph for a video %
\cite{rohrbach14gcpr,shin16icip,yu16cvpr}. Recently, \cite{yu2018fine} propose a fine-grained video captioning model for generating detailed sports narratives, and \cite{xiong2018move} propose the Move Forward and Tell approach, which localizes events and progressively decides when to generate the next sentence. This is related to the task of dense captioning \cite{krishna2017dense}, where videos are annotated with multiple localized sentences but the task does not require to produce a single coherent paragraph for the video.

\myparagraph{Reinforcement learning for caption generation.}
Most deep language generation models rely on Cross-Entropy loss and during training are given a previous ground-truth word. This is known to cause an exposure bias \cite{ranzato2015sequence}, as at test time the models need to condition on the predicted words. To overcome this issue, a number of reinforcement learning (RL) actor-critic \cite{konda2000actor} approaches have been proposed \cite{ren2017deep,rennie2017self,zhang2017actor}. \cite{liu2017improved} propose a policy gradient optimization method to directly optimize for language metrics, like CIDEr \cite{cider}, using Monte Carlo rollouts.
\cite{rennie2017self} propose a Self-Critical Sequence Training (SCST) method based on REINFORCE \cite{williams1992simple}, and instead of estimating a baseline, use the test-time inference algorithm (greedy decoding).

Recent works adopt similar techniques to video description. \cite{pasunuru2017reinforced} extend the approach of \cite{ranzato2015sequence} by using a mixed loss (both cross-entropy and RL) and correcting CIDEr with an entailment penalty.
\cite{wang2018video} propose a hierarchical reinforcement learning approach, where a Manager generates sub-goals, a Worker performs low-level actions, and a Critic determines whether the goal is achieved. Finally, \cite{li2018end} propose a multitask RL approach, built off \cite{rennie2017self}, with an additional attribute prediction loss. %

\myparagraph{GANs for caption generation.}
Instead of optimizing for hand-designed metrics, some recent works aim to learn what the ``good'' captions should be like using adversarial training. The first works to apply Generative Adversarial Networks (GANs) \cite{goodfellow2014generative} to image captioning are \cite{shetty2017speaking} and \cite{dai2017towards}. \cite{shetty2017speaking} train a discriminator to distinguish natural human captions from fake generated captions, focusing on caption diversity and image relevance. To sample captions they rely on Gumbel-Softmax approximation \cite{jang2016categorical}. \cite{dai2017towards} instead rely on policy gradient, and their discriminator focuses on caption naturalness and image relevance.
Some works have applied adversarial learning to generate paragraph descriptions for images/image sequences. \cite{liang2017recurrent} propose a joint training approach which incorporates multi-level adversarial discriminators, one for sentence level and another for coherent topic transition at a paragraph level. \cite{wang2018no} rely on adversarial reward learning to train a visual storytelling policy. \cite{wang2018show} use a multi-modal discriminator and a paragraph level language-style discriminator for their adversarial training. %
Their multi-modal discriminator resembles the standard discriminator design of \cite{dai2017towards,shetty2017speaking}. In contrast, we decouple the multi-modal discriminator into two specialized discriminators, Visual and Language, and use a Pairwise discriminator for sentence pairs' coherence.  Importantly, none of these works rely on their trained discriminators during inference.

Two recent image captioning works propose using discriminator scores instead of language metrics in the SCST model \cite{chen2018improving,melnyk2018improved}. We implement a GAN baseline based on this idea, and compare it to our approach.

\myparagraph{Caption sampling and re-scoring.} 
A few prior works explore caption sampling and re-scoring during inference \cite{andreas2016reasoning,hendricks2018grounding,vedantam2017context}. Specifically, \cite{hendricks2018grounding} aim to obtain more image-grounded bird explanations, while \cite{andreas2016reasoning,vedantam2017context} aim to generate discriminative captions for a given distractor image. While our approach is similar, our goal is different, as we work with video rather than images, and aim to improve multi-sentence description with respect to multiple properties.

\myparagraph{Alternatives to automatic metrics.} %
There is a growing interest in alternative ways of measuring the description quality, than \eg \cite{bleu,meteor,cider}. \cite{cui2018learning} train a general critic network to learn to score captions, providing various types of corrupted captions as negatives. \cite{sharif2018learning} use a composite metric, a classifier trained on the automatic scores as input. In contrast, we do not aim to build a general evaluation tool, but propose to improve the video description quality with our Adversarial Inference for a given generator.

\section{\label{sec:approach} Generation with Adversarial Inference}

\begin{figure*}[t]
    \centering
    \includegraphics[width=\linewidth]{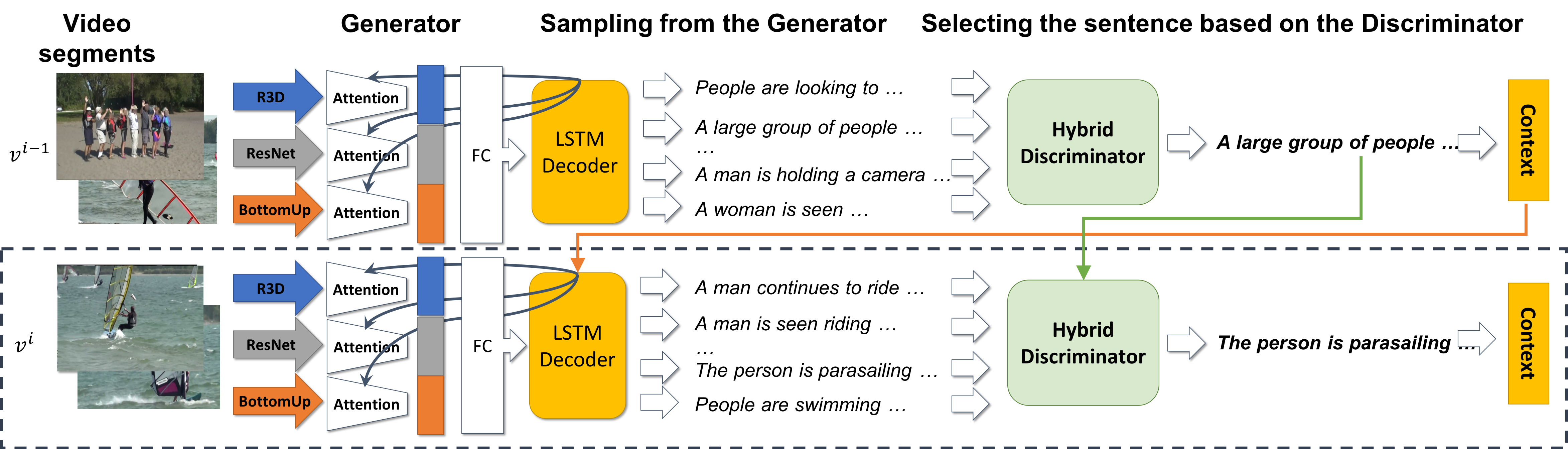}
    \caption{The overview of our Adversarial Inference approach. The Generator progressively samples candidate sentences for each clip, using the previous sentence as context. The Hybrid Discriminator scores the candidate sentences, and chooses the best one based on its visual relevance, linguistic characteristics and coherence to the previous sentence (details in Figure~\ref{fig:discriminators}).}
    \label{fig:overview}
\end{figure*}

In this section, we present our approach to multi-sentence description generation based on our \emph{Adversarial Inference} method. We first introduce our baseline generator $G$ and then discuss our discriminator $D$. The task of $D$ is to score the descriptions generated by $G$ for a given video. This includes, among others, to measure whether the multi-sentence descriptions are (1) correct with respect to the video, (2) fluent within individual sentences, and (3) form a coherent story across sentences. Instead of assigning all three tasks to a \emph{single} discriminator, we propose to compose $D$ out of three separate discriminators, each focusing on one of the above tasks. We denote this design a \emph{hybrid} discriminator (see Figure~\ref{fig:discriminators}).

While prior works mostly rely on discriminators for joint adversarial training \cite{dai2017towards,shetty2017speaking}, we argue that using them during inference is a more robust way of improving over the original generator. 
In our \emph{Adversarial Inference}, the pre-trained generator $G$ presents $D$ with the sentence candidates by sampling from its probability distribution. In its turn, our \emph{hybrid} discriminator $D$ selects the best sentence relying on the combination of its sub-discriminators. The overview of our approach is shown in Figure~\ref{fig:overview}. 

\subsection{Baseline Multi-Sentence Generator: $G$}

Given $L$ clips $[v^1, v^2, ..., v^L]$ from a video $v$, the task of $G$ is to generate $L$ sentences $[s^1, s^2, ..., s^L]$, where each sentence $s^i$ matches the content of the corresponding clip $v^i$. As the clips belong to the same video and are thus contextually dependent, %
our goal is to not only generate a sentence that matches its visual content, but to obtain a coherent and diverse sequence of sentences, \ie a natural paragraph.

Our generator follows a standard LSTM decoder \cite{donahue15cvpr,hochreiter1997long} to generate individual sentences $s^i$ with encoded representation of $v^i$ as our visual context. Typically, for each step $m$, the LSTM hidden state $h^i_m$ expects an input vector that encodes the visual features from $v^i$ as well as the previous word $w^i_{m-1}$. For our visual context, we use motion, RGB images, and object detections as features for each video clip, and follow the settings from \cite{wang2016temporal, xiong2018move} to obtain a single vector representation of each feature using a temporal attention mechanism \cite{yao2015iccv}\footnote{For details, please, see the supplemental material.}. The three vectors are concatenated to get the visual input ${\bar{v}^i}_m$. To encourage coherence among consecutive sentences, we additionally append the last hidden state of the previous sentence $h^{i-1}$ as input to the LSTM decoder \cite{gella2018dataset, xiong2018move}. The final input to the LSTM decoder for clip $v^i$ at time step $m$ is defined as follows: 
\begin{equation}
\begin{aligned} 
h^i_m = LSTM({\bar{v}^i}_m, w^i_{m-1}, h^{i-1}), \\
\text{with} \quad h^0 = 0,
\end{aligned}
\label{eq:gen}
\end{equation}

We follow the standard Maximum Likelihood Estimation (MLE) training for $G$, \ie we maximize the likelihood of each word $w^i_m$ given the current LSTM hidden state $h^i_m$. %

\subsection{Discriminator: $D$}
The task of a discriminator $D$ is to score a sentence $s$ w.r.t.~a video $v$ as $D(s|v) \in (0,1)$, where 1 indicates a positive match, while 0 is a negative match. Most prior works that perform adversarial training for image captioning \cite{chen2018improving,dai2017towards,melnyk2018improved,shetty2017speaking}, rely on the following ``single discriminator'' design. $D$ is trained to distinguish human ground-truth sentences as positives vs.~sentences generated by $G$ and mismatched ground truth sentences (from a different video) as negatives. The latter  aim to direct the discriminator's attention to the sentences' visual relevance. %

For a given generator $G$, the discriminator $D$ is trained with the following objective: 

\begin{equation}
\max \frac{1}{N}\sum_{j=1}^N{L_D(v^j)},
\end{equation}

where N is the number of training videos. For a video $v^j$ a respective term is defined as:

\begin{equation}
\begin{aligned}
L_D(v^j) = \mathbb{E}_{s\in S_{v^j}}[\text{log}(D(s|v^j))] \quad + \\
\mu \cdot \mathbb{E}_{s\in S_G}[\text{log}(1-D(s|v^j))] \quad + \\
\nu \cdot \mathbb{E}_{s\in S_{\setminus v^j}}[\text{log}(1-D(s|v^j))],
\end{aligned}
\label{eq:dis}
\end{equation}

where $S_{v^j}$ is the set of ground truth descriptions for $v^j$, $S_G$ are generated samples from $G$, $S_{\setminus v^j}$ are ground truth descriptions from \emph{other} videos, $\mu, \nu$ are hyper-parameters.

\subsubsection{Hybrid Discriminator}
In the ``single discriminator'' design, the discriminator is given multiple tasks at once, \ie to detect generated ``fakes'', which requires looking at linguistic characteristics, such as diversity or language structure, as well the mismatched ``fakes'', which requires looking at sentence semantics and relate it to the visual features. Moreover, for multi-sentence description, we would also like to detect cases where a sentence is inconsistent or redundant to a previous sentence.

\begin{figure}[t]
    \centering
    \includegraphics[width=\linewidth]{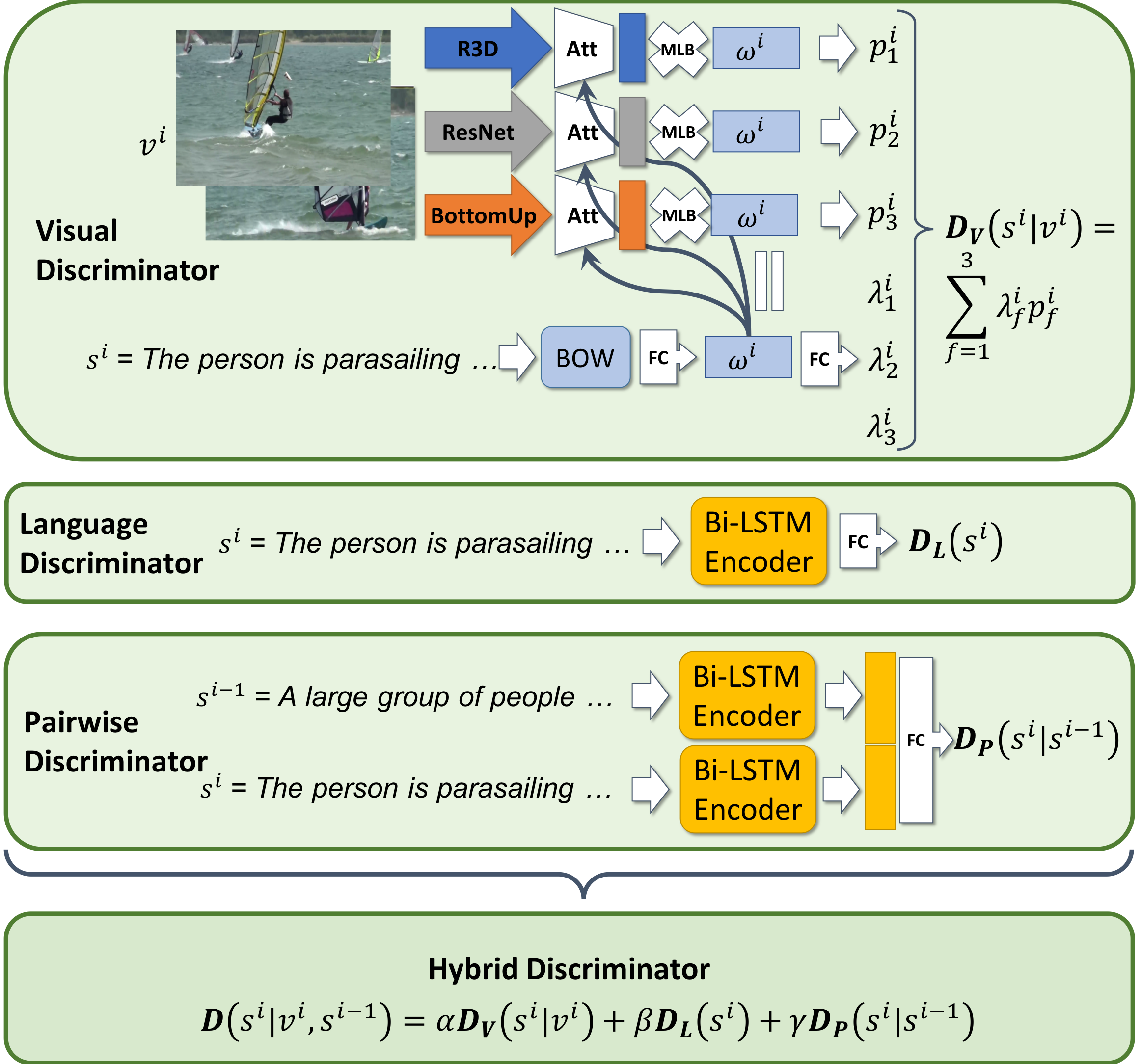}
    \caption{An overview of our Hybrid Discriminator. We score a sentence $s^i$ for a given video clip $v^i$ and a previous sentence $s^{i-1}$.}
    \label{fig:discriminators}
\end{figure}

To obtain these properties, we argue it is important to decouple the different tasks and allocate an individual discriminator for each one. In the following we introduce our visual, language and pairwise discriminators, which jointly constitute our \textit{hybrid discriminator} (see Figure~\ref{fig:discriminators}). We use the objective defined above for all three, however, the types of negatives vary by discriminator.

\paragraph{Visual Discriminator.} 
The v isual discriminator $D_V$ determines whether a sentence $s^i$ refers to concepts present in a video clip $v^i$, regardless of fluency and grammatical structure of the sentence. We believe that as the pre-trained generator already produces video relevant sentences, we should not include the generated samples as negatives for $D_V$. Instead, we use the mismatched ground truth as well as mismatched generated sentences as our two types of negatives. While randomly mismatched negatives may be easier to distinguish, hard negatives, \eg sentences from videos with the same activity as a given video, require stronger visual discriminative abilities. To improve our discriminator, we introduce such hard negatives, after training $D_V$ for 2 epochs.

Note, that if we use an LSTM to encode our sentence inputs to $D_V$, it may exploit the language characteristics to distinguish the generated mismatched sentences, instead of looking at their semantics. To mitigate this issue, we replace the LSTM encoding with a bag of words (BOW) representation, \ie each sentence is represented as a vocabulary-sized binary vector. The BOW is further embedded via a linear layer, and thus we obtain our final sentence encoding $\omega^i$. %

Similar to $G$, $D_V$ also considers multiple visual features, \ie we aggregate features from different misaligned modalities (video, image, objects). We individually encode each feature $f$ using temporal attention based on the entire sentence representation $\omega^i$. 
The obtained vector representations $\hat{v}^i_f$ are then fused with the sentence representation $\omega^i$, using Multimodal Low-rank Bilinear pooling (MLB) \cite{mlb2017}, which is known to be effective in tasks like multi-modal retrieval or VQA. The score for visual feature $f$ and sentence representation $\omega^i$ is obtained as follows:
 \begin{equation}
     p^i_f = \sigma(\tanh(U^T\hat{v}^i_f) \odot \tanh(V^T\omega^i)),
 \end{equation} 
where $\sigma$ is a sigmoid, producing values in $(0, 1)$, $\odot$ is the Hadamard product, $U$, $V$ are linear layers. 
Instead of concatenating features $\hat{v}^i_f$ as done in the generator, here we determine the scores $p^i_f$ between the sentence and each modality, and learn to weigh them adaptively based on the sentence. The intuition is that some sentences are more likely to require video features (``a man is jumping''), while others may require \eg object features (``a man is wearing a red shirt''). Following \cite{miech2018learning}, we assign weights $\lambda^i_f$ to each modality based on the sentence representation $\omega^i$:

\begin{equation}
 \lambda^i_f = \frac{e^{a_f^T \omega^i}}{\sum_{j} e^{ a_j^T \omega^i}},
\end{equation}
 
where $a_j$ are learned parameters. 
Finally, the $D_V$ score is the sum of the scores $p^i_f$ weighted by $\lambda^i_f$: 

\begin{equation}
 D_V(s^i | v^i) = \sum_f{\lambda^i_f p^i_f}.
\end{equation}

\paragraph{Language Discriminator.}
Language discriminator $D_L$ focuses on language structure of an individual sentence $s^i$, independent of its visual relevance. Here we want to ensure fluency as well as diversity of sentence structure that is lacking in $G$. The ActivityNet Captions \cite{krishna2017dense} dataset, that we experiment with, has long (over 13 words on average) and diverse descriptions with varied grammatical structures. 
In initial experiments we observed that a simple discriminator is able to point out a obvious mismatches based on diversity of the real vs. fake sentences, but fails to capture fluency or repeating N-grams. To address this, in addition to generated sentences from $G$, $D_L$ is given negative inputs with a mixture of randomly shuffled words or with repeated phrases within a sentence.

To obtain a $D_L$ score, we encode a sentence $s^i$ with a bidirectional LSTM, concatenate both last hidden states, denoted as $\bar h^{i}$, followed by a fully connected layer and a sigmoid layer: 
    \begin{equation}
        D_L(s^i) = \sigma(W_L\bar h^i+b_L).
    \end{equation}

\paragraph{Pairwise Discriminator.}
Pairwise discriminator $D_P$ evaluates whether two consecutive sentences $s^{i-1}$ and $s^i$ are coherent yet diverse in content. Specifically, $D_P$ scores $s^i$ based on $s^{i-1}$. To ensure coherence, we include ``shuffled'' sentences as negatives, \ie the order of sentences in a paragraph is randomly changed. We also design negatives with a pair of identical sentences $(s^i=s^{i-1})$ and optionally cutting off the endings (\eg ``a person enters and takes a chair'' and ``a person enters'') to avoid repeating contents. 

Similar to $D_L$ above, we encode both sentences with a bidirectional LSTM and obtain $\bar h^{i-1}$ and $\bar h^{i}$. We concatenate the two vectors and compute the $D_P$ score as follows:

    \begin{equation}
        D_P(s^i | s^{i-1}) = \sigma(W_P[\bar h^{i-1}, \bar h^i]+b_P).
    \end{equation}
    
Note, that the first sentence of a video description paragraph is not assigned a pairwise score, as there is no previous sentence. 

\subsection{Adversarial Inference}

In adversarial training for caption generation, $G$ and $D$ are first pre-trained and then jointly updated, where the discriminator improves the generator by providing feedback to the quality of sampled sentences. To deal with the issue of non-differentiable discrete sampling in joint training, several solutions have been proposed, such as Reinforcement Learning with variants of policy gradient methods or Gumbel softmax relaxation \cite{chen2018improving,dai2017towards,shetty2017speaking}. 
While certain improvement has been shown, as we discussed in Section~\ref{sec:intro}, GAN training can be very unstable. 

Motivated by the difficulties of joint training, we present our \textit{Adversarial Inference} method, which uses the discriminator $D$ during inference of the generator $G$. We show that our approach outperforms a jointly trained GAN model, most importantly, in human evaluation (see Section~\ref{sec:experiments}).
During inference, the generator typically uses greedy max decoding or beam search to generate a sentence based on the maximum probability of each word. One alternative to this is sampling sentences based on log probability \cite{donahue17pami}. Instead, we use our Hybrid Discriminator to score the sampled sentences. Note, that we generate sentences \emph{progressively}, \ie we provide the hidden state representation of the previous best sentence as context to sample the next sentence (see Figure~\ref{fig:overview}). Formally, for a video clip $v^i$, a previous best sentence $s^{i-1}_*$ and $K$ sampled sentences ${s^i_1, s^i_2, ... s^i_K}$ from the generator $G$, the scores from our hybrid discriminator can be used to compare the sentences and select the best one:
\begin{equation}
    s^i_* = s^i_{\text{argmax}_{j=1..K}{D(s^i_j|v^i,s^{i-1}_*))}},
\end{equation}

where $s^i_j$ is the $j^\text{th}$ sampled sentence. The final discriminator score is defined as: 
\begin{equation}
\begin{aligned}
    D(s^i_j|v^i,s^{i-1}_*) = \alpha \cdot D_V(s^i_j|v^i) \quad + \\ \beta \cdot D_L(s^i_j) + \gamma \cdot  D_P(s^i_j|s^{i-1}_*),
\end{aligned}
\end{equation}
where $\alpha, \beta, \gamma$ are hyper-parameters. %

\section{\label{sec:experiments} Experiments}

\newcommand{\midruleStatsNew}{\cmidrule(rr){1-1} \cmidrule(rr){2-4} \cmidrule(rr){5-6} \cmidrule(rl){7-7} \cmidrule(rl){8-10}}

\begin {table*}[t]
\begin{center}
\small
\begin{tabular}{@{}l@{\ }c@{\ \ \ }c@{\ \ \ }c@{\ \ \ }c@{\ \ \ }c@{\ \ \ }c@{\ \ \ }c@{\ }c@{\ }c@{\ }c@{\ }c@{\ }c@{}}
\toprule
\textbf{} & \multicolumn{3}{c}{\textbf{Per video}} & \multicolumn{2}{c}{\textbf{Overall}} & \multicolumn{1}{c}{\textbf{Per act.}} & \multicolumn{3}{c}{\textbf{Per video}}\\
\textbf{Method} & \textbf{METEOR} & \textbf{BLEU@4} & \textbf{CIDEr-D} %
& \textbf{Vocab} & \textbf{Sent} & \textbf{RE-4 $\downarrow$} & \textbf{Div-1 $\uparrow$} & \textbf{Div-2 $\uparrow$} & \textbf{RE-4 $\downarrow$}\\
\textbf{} & \textbf{} & \textbf{} & \textbf{} & \textbf{Size} & \textbf{Length} &  &  &  & \\
\midruleStatsNew
MLE & 16.70  & 9.95 & 20.32 %
& 1749 & 13.83 & 0.38 & 0.55 & 0.74 & 0.08 \\ 
\midruleStatsNew
GAN w/o CE & 16.49 & 9.76 & 20.24  %
& 2174 & 13.67 & 0.35 & 0.56 & 0.74  & 0.07  \\
GAN & 16.69 & 10.02 &  21.07 %
& 1930 & 13.60 & 0.36 & 0.56 & 0.74  & 0.07 \\
\midruleStatsNew
SCST & 15.80 & 10.82 & 20.89 %
& 941 & 12.13 & 0.52 & 0.47 & 0.65 & 0.11 \\
\midruleStatsNew
MLE + BS3 & 16.22 & 10.79 & 21.81 %
& 1374 & 12.92 & 0.48 & 0.55 & 0.71 & 0.11 \\
MLE + LP & {17.51} & 8.70 & 12.23 %
& 1601 & 18.68 & 0.48 & 0.48 & 0.69 & 0.12 \\
\midruleStatsNew
MLE + SingleDis & 16.29 & 9.25 & 18.17  %
& 2291 & 13.98 & 0.37 & 0.59 & 0.75 & 0.07 \\
MLE + SingleDis w/ Pair & 16.16 & 9.32 & 18.72 & 2375 & 13.75 & 0.37 & {0.60} & 0.77 & 0.06  \\
\midruleStatsNew
(Ours) MLE + HybridDis w/o Vis & 16.33 & 8.92 & 17.29 & 2462 & 14.43 & 0.34 & 0.59 & 0.76 & 0.06  \\
(Ours) MLE + HybridDis w/o Lang & 16.44 & 9.37 & 19.44 & {2697} & 13.77 & {0.30} & 0.59 & {0.78} & {0.05} \\
(Ours) MLE + HybridDis w/o Pair & 16.60 & 9.56 & 19.39 %
& 2390 & 13.86 & 0.32 & 0.58 & 0.76 & 0.06  \\
(Ours) MLE + HybridDis & 16.48 & 9.91 & 20.60 %
& 2346 & 13.38 & 0.32 & 0.59 & 0.77 &  0.06 \\
\midruleStatsNew
Human & - & - & -  & 8352 & 14.27 & 0.04 & 0.71 & 0.85 & 0.01 \\
\midrule
\multicolumn{10}{c}{\textbf{SoTA models}} \\
\midrule
VideoStory \cite{gella2018dataset}  & 16.26  & 7.66  & 14.53  %
& 1269 & 16.73 & 0.37 & 0.51 & 0.72  & 0.09 \\
Transformer \cite{zhou2018end}  & 16.15 & 10.29 & 21.72 %
& 1819 & 12.42 & 0.34 & 0.53 & 0.73  & 0.07 \\
MoveForwardTell \cite{xiong2018move}  & 14.67  & 10.03  & 19.49  %
& 1926 & 11.46 & 0.53 & 0.55 & 0.66 & 0.18 \\
\bottomrule
\end{tabular}
\end{center}
\vspace{-0.5cm}
\caption{Comparison to video description baselines and SoTA models. Statistics over generated descriptions include N-gram Diversity (Div-1,2, higher better) and Repetition (RE-4, lower better) per video and per activity. See Section~\ref{sec:results} for details.} 
\label{tab:ablations}
\end {table*}

We benchmark our approach for multi-sentence video description on the ActivityNet Captions dataset \cite{krishna2017dense} and compare our Adversarial Inference to GAN and other baselines, as well as to state-of-the-art models.

\subsection{Experimental Setup}
\myparagraph{Dataset.}
The ActivityNet Captions dataset contains 10,009 videos for training and 4,917 videos for validation with two reference descriptions for each\footnote{The two references are not aligned to the same time intervals, and even may have a different number of sentences.}. Similar to prior work \cite{zhou2018end,gella2018dataset}, %
we use the validation videos with the \nth{2} reference for development, while the \nth{1} reference is used for evaluation. While the original task defined on ActivityNet Captions involves both event localization and description, we run our experiments with ground truth video intervals. Our goal is to show that our approach leads to more correct, diverse and coherent multi-sentence video descriptions.%

\myparagraph{Visual Processing.} Each video clip is encoded with 2048-dim ResNet-152 features \cite{he2016deep} pre-trained on ImageNet \cite{deng2009imagenet} (denoted as ResNet) and 8192-dim ResNext-101 features \cite{hara2018can} pre-trained on the Kinetics dataset \cite{kay2017kinetics} (denoted as R3D). We extract both ResNet and R3D features at every 16 frames and use a temporal resolution of 16 frames for R3D. %
The features are uniformly divided into 10 segments as in \cite{wang2016temporal, xiong2018move},
and mean pooled within each segment to represent the clip as 10 sequential features. We also run the Faster R-CNN detector \cite{ren2015faster} from \cite{anderson2018bottom} trained on Visual Genome \cite{krishna2017visual}, on 3 frames (at the beginning, middle and end of a clip) and detect top 16 objects per frame.%
We encode the predicted object labels with bag of words weighted by detection confidences (denoted as BottomUp). %
Thus, a visual representation for each clip consists of 10 R3D features, 10 ResNet features, and 3 BottomUp features.

\myparagraph{Language Processing.} The sentences are ``cut'' at a maximum length of 30 words. 
The LSTM cells' dimensionality is fixed to 512. The discriminators' word embeddings are initialized with 300-dim Glove embeddings \cite{pennington2014glove}.

\myparagraph{Training and Inference.} 
 We train the generator and discriminators with cross entropy objectives using the ADAM optimizer \cite{kingma2014adam} with a learning rate of $5e^{-4}$. One batch consists of multiple clips and captions from the same video, and the batch size is fixed to 16 when training all models. The weights for all the discriminators' negative inputs ($\mu$, $\nu$ in Eq.~\ref{eq:dis}), are set to 0.5. The weights for our hybrid discriminator are set as $\alpha$ = 0.8, $\beta$ = 0.2, $\gamma$ = 1.0. Sampling temperature during discriminator training is 1.0; during inference we sample $K = 100$ sentences with temperature 0.2. When training the discriminators, a specific type of a negative example is randomly chosen for a video, \ie a batch consists of a combination of different types of negatives. %

\myparagraph{Baselines and SoTA.}
We compare our Adversarial Inference (denoted MLE+HybridDis) to: our baseline generator (MLE); multiple inference procedures, \ie beam search with size 3 (MLE+BS3), sampling with log probabilities (MLE+LP) and inference with the single discriminator (MLE+SingleDis); Self Critical Sequence Tranining \cite{rennie2017self} which optimizes for CIDEr (SCST); GAN models built off \cite{chen2018improving,melnyk2018improved} with a single discriminator\footnote{We have tried incorporating our hybrid discriminator in GAN training, however, we have not observed a large difference, likely due to a large space of training hyper-parameters which is challenging to explore.}, with and without a cross entropy (CE) loss (GAN, GAN w/o CE). Finally, we also compare to the following state-of-the-art methods: Transformer \cite{zhou2018end}, VideoStory \cite{gella2018dataset} and MoveForwardTell \cite{xiong2018move}, whose predictions we obtained from the authors.

\subsection{Results \label{sec:results}}

\myparagraph{Automatic Evaluation.}
Following \cite{xiong2018move}, we conduct our evaluation at paragraph-level. We include standard metrics, \ie METEOR \cite{meteor}, BLEU@4 \cite{bleu} and CIDEr-D \cite{cider}. However, these alone are not sufficient to get a holistic view of the description quality, since the scores fail to capture content diversity or detect repetition of phrases and sentence structures. To see if our approach improves on these properties, we report Div-1 and Div-2 scores \cite{shetty2017speaking}, that measure a ratio of unique N-grams (N=1,2) to the total number of words, and RE-4 \cite{xiong2018move}, that captures a degree of N-gram repetition (N=4) in a description\footnote{For Div-1,2 higher is better, while for RE-4 lower is better.}. We compute these scores at video (paragraph) level, and report the average score over all videos. 
Finally, we want to capture the degree of ``discriminativeness'' among the descriptions of videos with similar content. ActivitiyNet~\cite{caba2015activitynet} includes 200 activity labels, and the videos with the same activity have similar visual content. We thus also report RE-4 per activity by combining all sentences associated with each activity, and averaging the score over all activities.

We compare our model to baselines in Table~\ref{tab:ablations} (top). The best performing models in standard metrics do not include our adversarial inference procedure nor the jointly trained GAN models. This is somewhat expected, as prior work shows that adversarial training does worse in these metrics than the MLE baseline \cite{dai2017towards,shetty2017speaking}. We note that adding a CE loss benefits GAN training, leading to more fluent descriptions (GAN w/o CE vs. GAN). We also observe that the METEOR score, popular in video description literature, is strongly correlated with sentence length. 

We see that our Adversarial Inference leads to more diverse descriptions with less repetition than the baselines, including GANs. Our MLE+HybridDis model outperforms the MLE+SingleDis in every metric, supporting our hybrid discriminator design. Furthermore, MLE + SingleDis w/ Pair scores higher than the SingleDis but lower than our HybridDis. This shows that a \emph{decoupled} Visual discriminator is important for our task. Note that the SCST has the lowest diversity and highest repetition among all baselines. Our MLE+HybridDis model also improves over baselines in terms of repetition score ``per activity'', suggesting that it obtains more video relevant and less generic descriptions.

\begin{figure}[t]
\scriptsize
\begin{center}
\includegraphics[width=\linewidth]{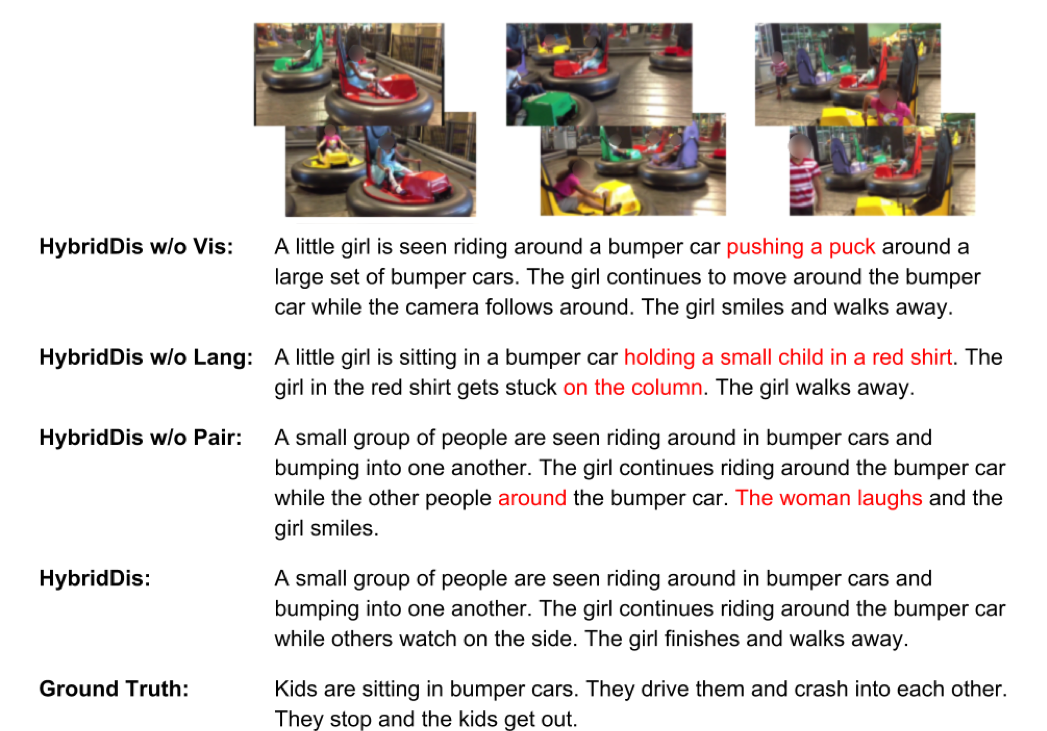} %
\caption{Comparison of ablated models vs. our full model (discussion in text). Content errors are highlighted in red.}
\label{fig:ablation_dis}
\end{center}
\end{figure}

To show the importance of all three discriminators, we provide ablation experiments by taking out each component, respectively (w/o Vis, w/o Lang, w/o Pair). Our HybridDis performs the worst when without its visual component and the combination of three discriminators outperforms each of the ablations on the standard metrics. %
In Figure~\ref{fig:ablation_dis}, we show a qualitative result obtained by the ablated models vs.~our full model. Removing the Visual discriminator leads to incorrect mention of  ``pushing a puck'', as the visual error is not penalized as needed. Model without the Language discriminator results in somewhat implausible constructs (``stuck in the column'') and incorrectly mentions ``holding a small child''. Removing the Pairwise discriminator leads to incoherently including a ``woman'' while missing the salient ending event (kids leaving).

\begin {table}[t]
\begin{center}
\small
\begin{tabular}{@{}l@{\ \ }c@{\ \ }c@{\ \ }c@{}}
\toprule
\textbf{Method} & \textbf{Better} & \textbf{Worse} & \textbf{Delta}\\
\textbf{} & \textbf{than MLE} & \textbf{than MLE} & \textbf{}\\
\midrule
SCST & 22.0 & 62.0 & -40.0 \\ 
GAN & 32.5 & 30.0 & +2.5 \\ 
\midrule
MLE + BS3 & 27.0 & 31.0 & -4.0 \\ 
MLE + LP & 32.5 & 34.0 & -1.5 \\ 
MLE + SingleDis & 29.0 & 30.0 & -1.0 \\
(Ours) MLE + HybridDis w/o Pair & 42.0 & 36.5 & +5.5 \\
(Ours) MLE + HybridDis & 38.0 & 31.5 & \textbf{+6.5} \\ 
\bottomrule
\end{tabular}
\end{center}
\caption {Human evaluation of multi-sentence video descriptions, see text for details.} \label{tab:human} 
\end {table}

\myparagraph{Human Evaluation.}
The most reliable way to evaluate the description quality is with human judges. We run our evaluation on Amazon Mechanical Turk (AMT)\footnote{\url{https://www.mturk.com}} with a set of 200 random videos. To make the task easier for humans we compare two systems at a time, rather than judging multiple systems at once. We design a set of experiments, where each system is being compared to the MLE baseline. The human judges can select that one description is better than another or that both as similar. We ask 3 human judges to score each pair of sentences, so that we can compute a majority vote (\ie at least 2 out of 3 agree on a judgment), see results in Table~\ref{tab:human}. Our proposed approach improves over all other inference procedures, as well as over GAN and SCST. We see that the GAN is rather competitive, but still overall not scored as high as our approach. Notably, SCST is scored rather low, which we attribute to its grammatical issues and high redundancy in the descriptions.

\myparagraph{Comparison to SoTA.}
We compare our approach to multiple state-of-the-art methods using the same automatic metrics as above. As can be seen from Table~\ref{tab:ablations} (bottom), 
our MLE + HybridDis model performs on par with the state-of-the-art on standard metrics and wins in diversity metrics. We provide a qualitative comparison to the state-of-the-art models in Figure~\ref{fig:teaser} and in the supplemental material.

\begin {table}[t]
\begin{center}
\small
\begin{tabular}{@{}l@{\ \ \ }cl@{\ \ \ }c@{}}
\toprule
\textbf{Method} & \textbf{Exact} & \textbf{Gender+}\\
\textbf{} & \textbf{word} & \textbf{plurality}\\
\midrule
VideoStory \cite{gella2018dataset}  &  44.9 & 64.1 \\
Transformer \cite{zhou2018end}  & 45.8 &  66.0 \\
MoveForwardTell \cite{xiong2018move}  & 42.6  &  64.1 \\
\midrule
MLE & 48.8 & 67.5 \\
SCST & 44.0 & 63.3 \\
GAN & 48.9 & 67.5 \\
(Ours) MLE + HybridDis & \textbf{49.1} & \textbf{67.9} \\
\bottomrule
\end{tabular}
\end{center}
\caption {Correctness of person-specific words, F1 score.}
\vspace{-0.5cm}
\label{tab:f1}
\end {table}

\myparagraph{Person Correctness.}
Most video descriptions in the ActivityNet Captions dataset discuss people and their actions. To get additional insights into correctness of the generated descriptions, we evaluate the ``person words'' correctness. Specifically, we compare (a) the exact person words (e.g. \emph{girl}, \emph{guys}) and (b) only gender with plurality (e.g. \emph{female-single}, \emph{male-plural}) between the references and the predicted descriptions, and report the \emph{F1} score in Table~\ref{tab:f1} (this is similar to \cite{rohrbach17cvpr}, who evaluate character correctness in movie descriptions). Interestingly, our MLE baseline already outperforms the state-of-the-art in terms of person correctness, likely due to the additional object-level features \cite{anderson2018bottom}. SCST leads to a significant decrease in person word correctness, while our Adversarial Inference improves it.

\section{\label{sec:conclusion} Conclusion}
The focus of prior work on video description generation has so far been on training better generators and improving the input representation. In contrast, in this work we advocate an orthogonal direction to improve the quality of video descriptions: We propose the concept \emph{Adversarial Inference} for video description where a trained discriminator selects the best from a set of sampled sentences. This allows to make the final decision on what is the best sample \emph{a posteriori} by relying on strong trained discriminators, which look at the video and the generated sentences to make a decision. %
More specifically, we introduce a \emph{hybrid discriminator} which consists of three individual experts: one for language, one for relating the sentence to the video, and one pairwise, across sentences.
In our experimental study, humans prefer sentences selected by our \emph{hybrid discriminator} used in \emph{Adversarial Inference} better than the default greedy decoding. Beam search, sampling with log probability as well as previous approaches to improve the generator (SCST and GAN) are judged not as good as our sentences. We include further qualitative results which demonstrate the strength of our approach in supplemental materials.

\paragraph{Acknowledgements.}
The work of Trevor Darrell and Anna Rohrbach was in part supported by the DARPA XAI program, the
Berkeley Artificial Intelligence Research (BAIR) Lab, and the Berkeley DeepDrive (BDD) Lab.

\clearpage
\section*{Supplemental Material}
\appendix

Here we provide implementation details for our approach and baseline models (Section~\ref{sec:impl_supp}), %
and include qualitative comparison of our approach to ablations, baselines and state-of-the-art methods (Section~\ref{sec:examp_supp}).

\section{Implementation Details \label{sec:impl_supp}}

\myparagraph{Processing the Visual Feature.}
First, we detail how we obtain the visual input ${\bar{v}^i}_m$ in Equation~\ref{eq:gen}. %
Unlike image captioning that relies on static features, video description requires a dynamic multimodal fusion over different visual features, such as e.g. a stream of RGBs and motion features. %
In addition to video and image-level features, we introduce object detections extracted for a subset of frames. Different features may be temporally misaligned (\ie extracted over  different sets of frames). We address this as  follows. Suppose, a visual feature $f$ extracted from $v^i$ is represented as a sequence of $T_f$ segments: $v^i_f = [v^i_{f,1}, v^i_{f,2}, ... v^i_{f,T_f}]$~\cite{wang2016temporal, xiong2018move}. The previous hidden state $h^i_{m-1}$ is used to predict temporal attention  \cite{yao2015iccv} over these segments, which then results in a single feature vector $\hat{v}^i_{m,f}$. We concatenate the resulting vectors from all features as our final visual input to the decoder: $\bar{v}^i_m = [\hat{v}^i_{m,1}, \hat{v}^i_{m,2}, ..., \hat{v}^i_{m,f}, ...]$.

\myparagraph{Self-Critical Sequence Training.}
Self-Critical Sequence Training \cite{rennie2017self} (SCST)\footnote{Our SCST model is based on the implementation of \url{https://github.com/ruotianluo/self-critical.pytorch}\label{code}} is a variant of REINFORCE \cite{williams1992simple} where the inference algorithm is used as a baseline. Suppose we have a generator model $G_\theta$ with parameters $\theta$; a complete sequence $x^s=(x^s_1, ... x^s_{T})$ is sampled using the probability distribution $p_\theta(x^s_t | x^s_1, ... x^s_{t-1})$ at each time step $t$. To reduce the variance during training and explore beyond the current best policy, SCST decodes another sequence $\hat{x}$ with the inference algorithm (greedy decoding) and aims to improve $x^s$ over $\hat{x}$ based on a reward $r$ such as a CIDER metric \cite{cider}. The gradient function for the model is calculated as: 
\begin{equation}
    \nabla_\theta{L_{G_\theta}(\theta)} = \sum_{t=1}^{T}{(r(x^s) - r(\hat{x})) \nabla_\theta \log {p_\theta(x^s_t | x^s_{1:t-1})} }.
\end{equation}

\myparagraph{GANs for Captioning.}
GANs for image captioning \cite{dai2017towards,shetty2017speaking} are typically trained with the following procedure due to their instability in early training stages: 1) pre-train the generator $G_\theta$ optimizing MLE objective, 2) pre-train discriminator $D_\eta$ by sampling sentences from pre-trained $G_\theta$, and 3) jointly update $G_\theta$ and $D_\eta$ iteratively with a different objective for $G_\theta$ to deal with non-differentiable sampling. Cross Entropy loss is used to pre-train $G_\theta$ and $D_\eta$, where $D_\eta$ is trained with negative samples as in Equation~\ref{eq:dis}, %
with $\mu = 0.5, \nu = 0.5$. After both $G_\theta$ and $D_\eta$ have been pre-trained, we follow \cite{chen2018improving,melnyk2018improved} and jointly train them using SCST but replacing reward $r$ with an output of a standard (``single'') discriminator $D_\eta(V,x^s)$, where $V$ is a given video segment and $x^s$ is a sampled description. We find that it is best to update $G_\theta$ for 5 steps for each update of $D_\eta$. The gradient for the above GAN model is:

\begin{equation}
\begin{aligned}
    \nabla_\theta{L_{G_\theta}(\theta)} = \sum_{t=1}^{T}{(D_\eta(V,x^s) - D_\eta(V,\hat{x})) \nabla_\theta \log {p_\theta(x^s_t | x^s_{1:t-1})} }.
    \label{eq:grad}
\end{aligned}
\end{equation}

\begin{figure*}[t]
\scriptsize
\begin{center}
\subfloat[]{{\includegraphics[width=0.5\linewidth]{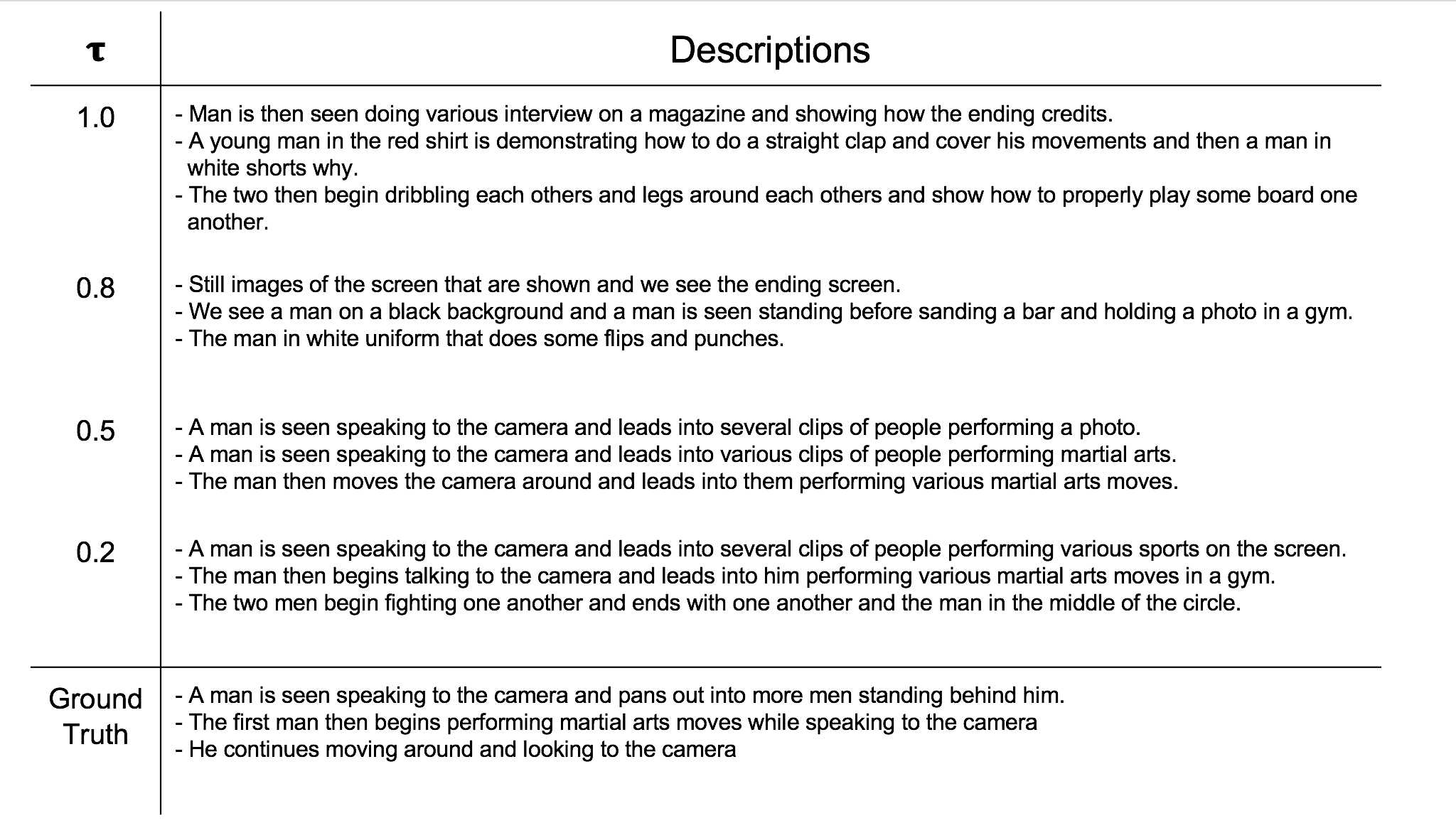} }}%
\subfloat[]{{\includegraphics[width=0.5\linewidth]{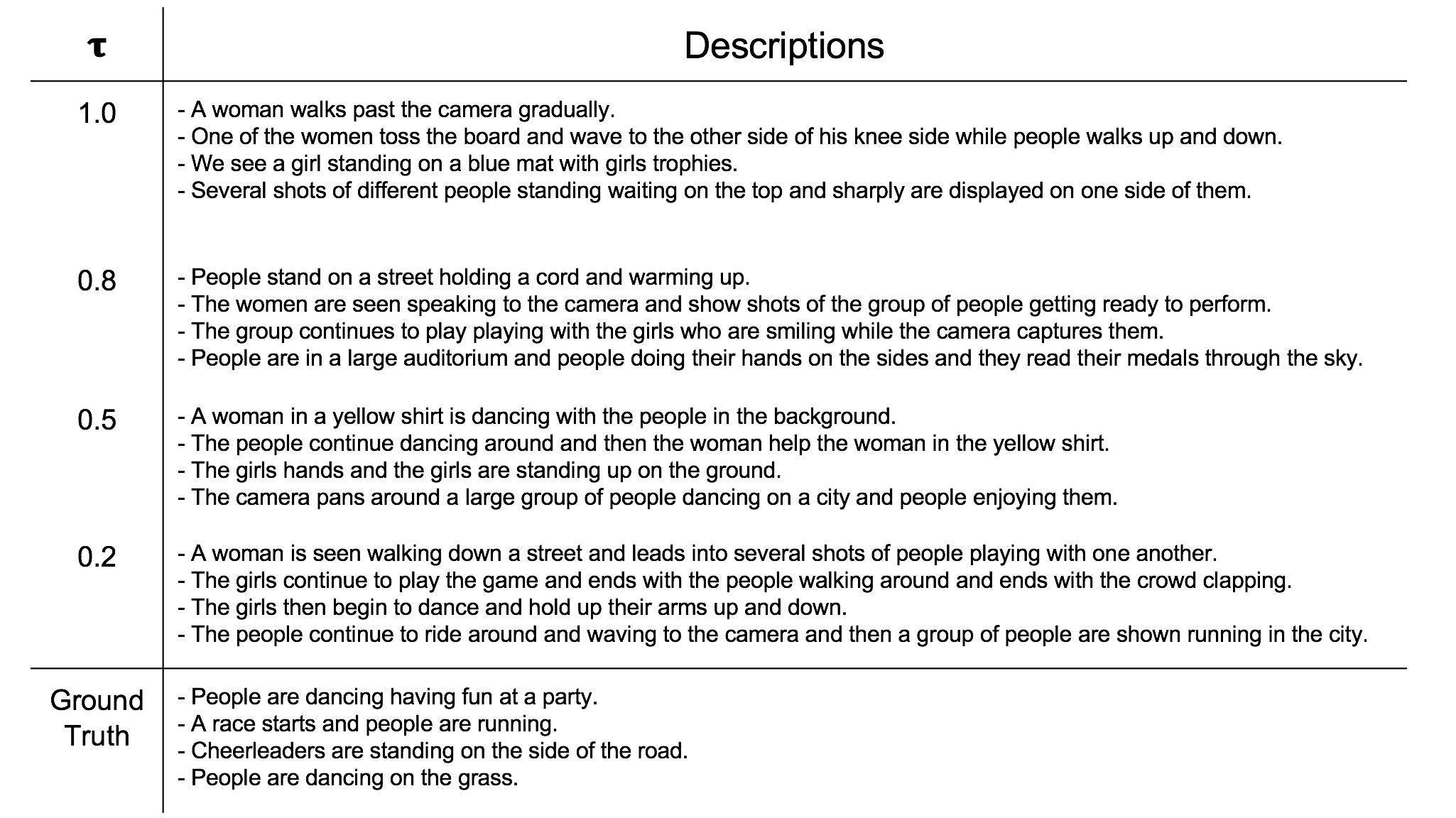} }}%
\caption{Sampling multi-sentence descriptions with different temperatures. The sentences are sampled from a pre-trained generator with temperatures $\{1.0,0.8,0.5,0.2\}$. Each sentence corresponds to a clip in a video. Note that higher temperatures tend to lead to more diverse vocabulary with the cost of decreased fluency.}
\label{fig:temp}
\end{center}
\end{figure*}

Due to instability of adversarial training, we additionally include a cross entropy (CE) loss that ensures that the generator will explore an output space in a more stable manner and maintain its language model \cite{pasunuru2017reinforced}. The final objective of $G_\theta$ is a mixed loss function, a weighted combination of Cross-Entropy Loss ($L_{\text{CE}}$) optimizing the maximum-likelihood training objective and Adversarial Loss ($L_{\text{GAN}}$) with its gradient function defined in Equation~\ref{eq:grad}:
\begin{equation}
L_{\text{MIX}} = \lambda L_{\text{GAN}} + (1-\lambda) L_{\text{CE}},
\end{equation}
where we use $\lambda = 0.995$.
We compare this mixed objective to not using the CE loss in Table~1 of the main paper.

\myparagraph{Adversarial Inference.}
Suppose each word $w_i$ in a vocabulary of size $K$ can be sampled with a  probability $p(w_i)$. One can additionally modify the probability distribution during sampling with a temperature parameter $\tau$:
\begin{equation}
p_\tau(w_i) = \frac{p(w_i)^{1/\tau}}{\sum_{j=1}^K {p(w_j)^{1/\tau}}}.
\label{eq:tau}
\end{equation}

Based on Equation~\ref{eq:tau}, $\tau = 1 $ is a default sampling procedure. Setting  %
$\tau < $ 1 shifts the distribution to favor larger probabilities, making the overall distribution more ``peaky''. %
We explore parameter $\tau$ for both discriminator training, $\tau_T$, and adversarial inference, $\tau_I$. 
We obtain more fluent captions by setting $\tau_I < 1$ during inference, however we find it is best to set $\tau_T = 1$ during discriminator training so that it learns to distinguish natural and fake descriptions. In our adversarial inference procedure, we sample $K = 100$ sentences with $\tau_I = 0.2$ for each for each video segment. %
One can see the effect of different temperatures during inference in Figure~\ref{fig:temp}.

\section{Qualitative Examples  \label{sec:examp_supp}}
In this section, we provide qualitative examples comparing our Adversarial Inference method to its ablations, other baselines and state-of-the-art models.

\begin{figure*}%
\scriptsize
\begin{center}
\subfloat[]{{\includegraphics[width=14cm]{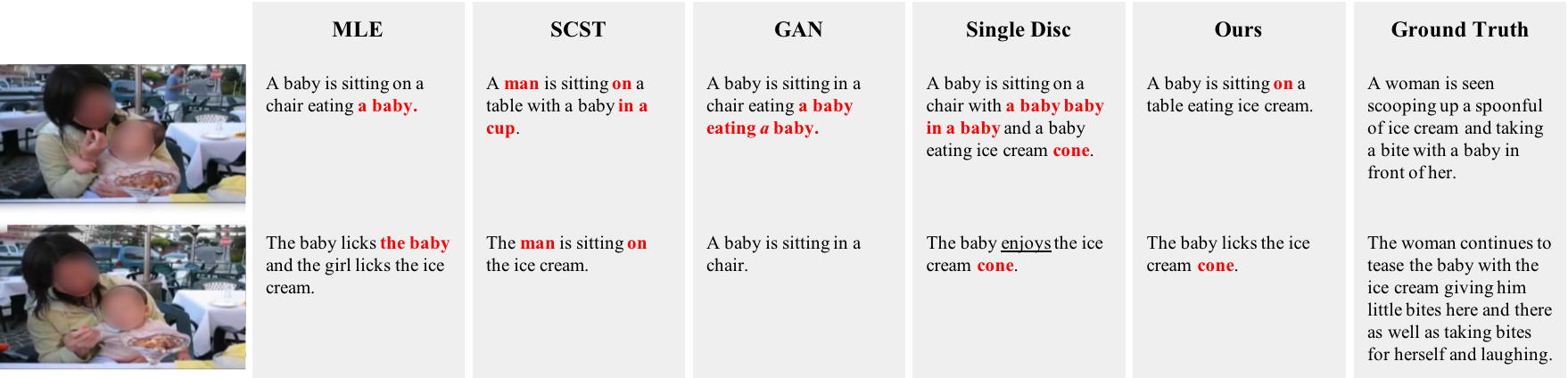} }}%
\quad
    \subfloat[]{{\includegraphics[width=14cm]{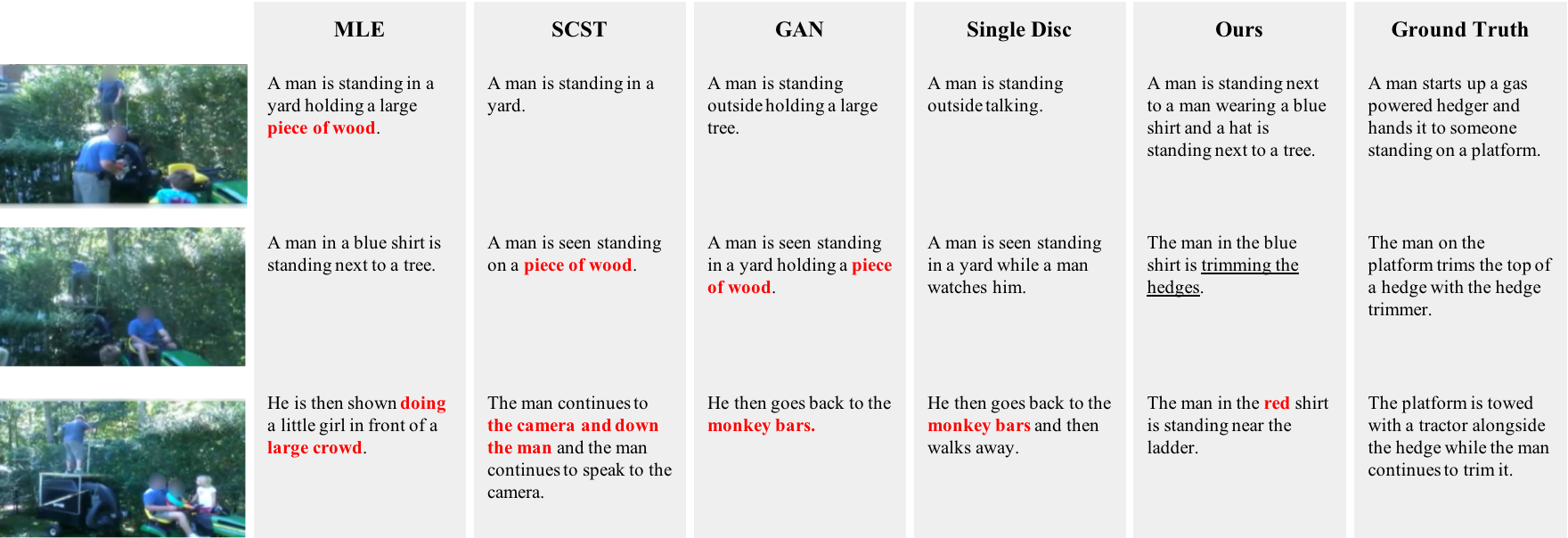} }}%
    
\quad
    \subfloat[]{{\includegraphics[width=14cm]{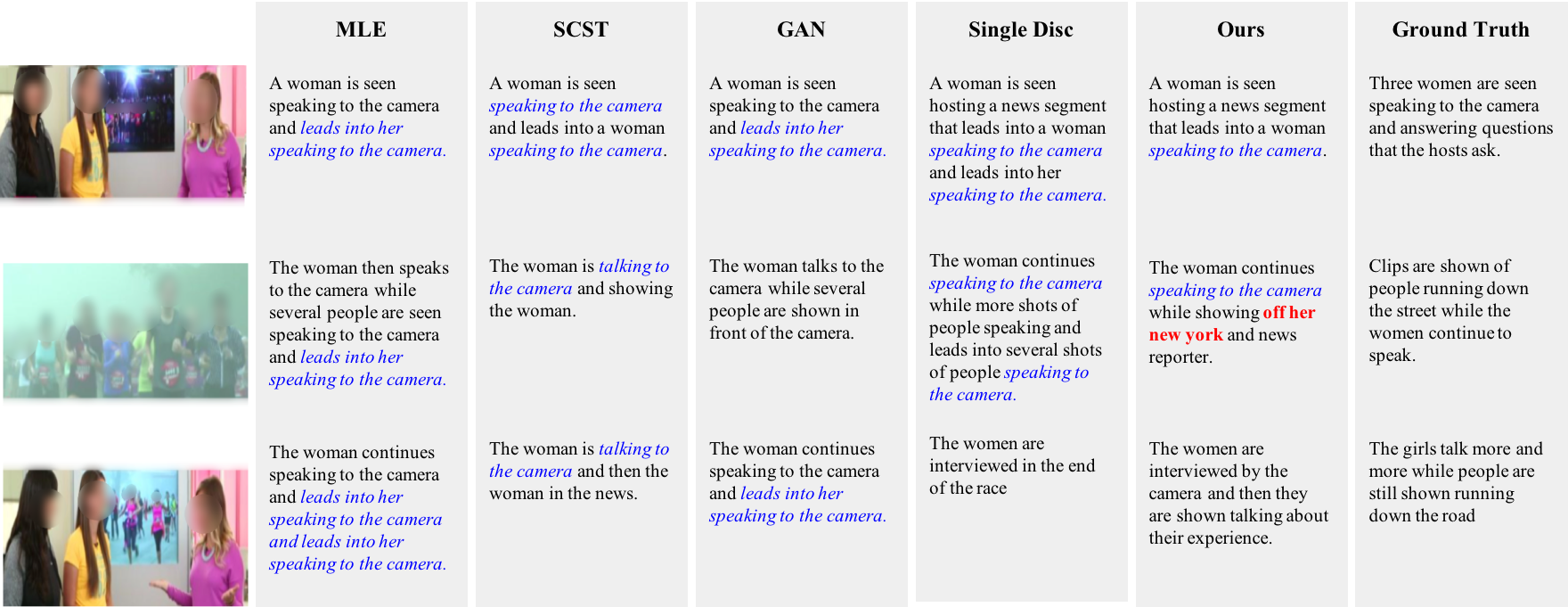} }}%
\caption{Comparison of our approach to MLE baseline, SCST, GAN, and Adversarial Inference with Single Discriminator. Red/bold indicates content errors, blue/italic indicates repetitive patterns.}
\label{fig:examples_ours}
\end{center}
\end{figure*}

\subsection{Comparison to Model Ablations and GAN}

Figure~\ref{fig:examples_ours} shows a few qualitative examples comparing ground truth descriptions to the ones generated by the following methods: MLE, SCST (with CIDEr), GAN, MLE+SingleDis (Single Disc), and our MLE+HybridDis (Ours). We highlight errors, \eg objects not present in video, in bold/red, and repeating phrases in italic/blue. Overall, our approach leads to more correct, more fluent, and less repetitive multi-sentence descriptions than the baselines. In (a), our prediction is preferable to all the baselines w.r.t. the sentence fluency. While all models recognize the presence of a baby and a person eating an ice cream, the baselines fail to describe the scene in a coherent way, but our approach summarizes the visual information correctly. Our model also generates more diverse descriptions specific to what is happening in the video, often mentioning more interesting and informative words/phrases, such as ``trimming the hedges'' in (b) or ``their experience'' in (c). MLE and SCST mention less visually specific information, and generate more generic descriptions, such as ``holding a piece of wood''. In an attempt to explore diverse phrases, the single discriminator is more prone to hallucinating non-existing objects, \eg ``monkey bars'' in (b). Finally, our model outperforms the baselines in terms of lower redundancy across sentences. As seen in (c), our approach delivers more diverse content for each clip, while all others more frequently generate ``speaking/talking to the camera'', a very common phrase in the dataset.

\begin{figure*}[t]
\scriptsize
\begin{center}
\subfloat[]{{\includegraphics[width=13cm]{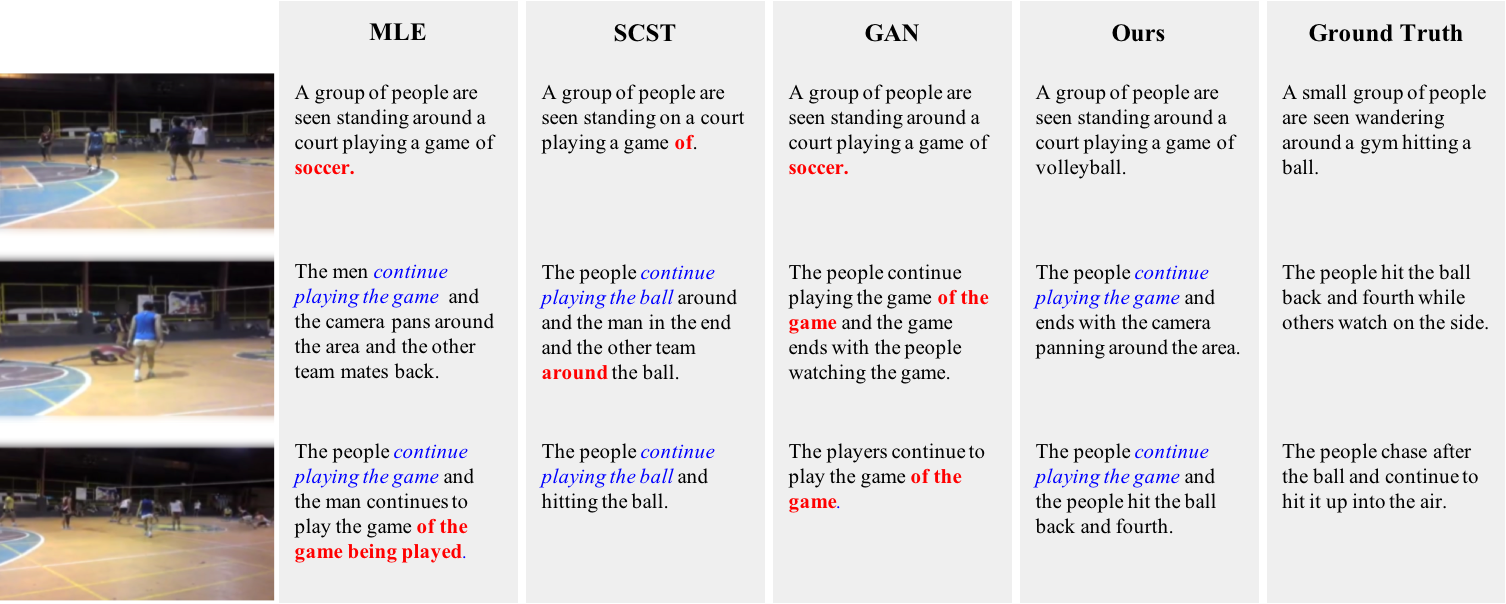} }}%
\quad
    \subfloat[]{{\includegraphics[width=13cm]{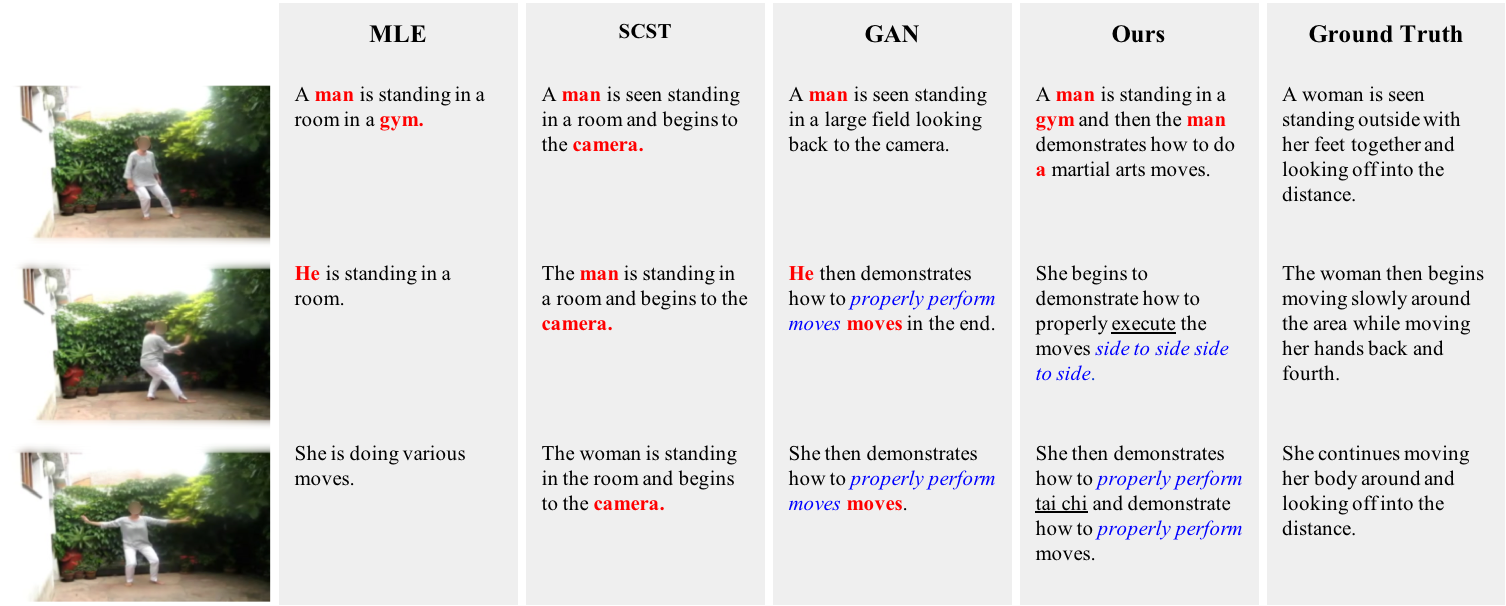} }}%
\quad
    \subfloat[]{{\includegraphics[width=13cm]{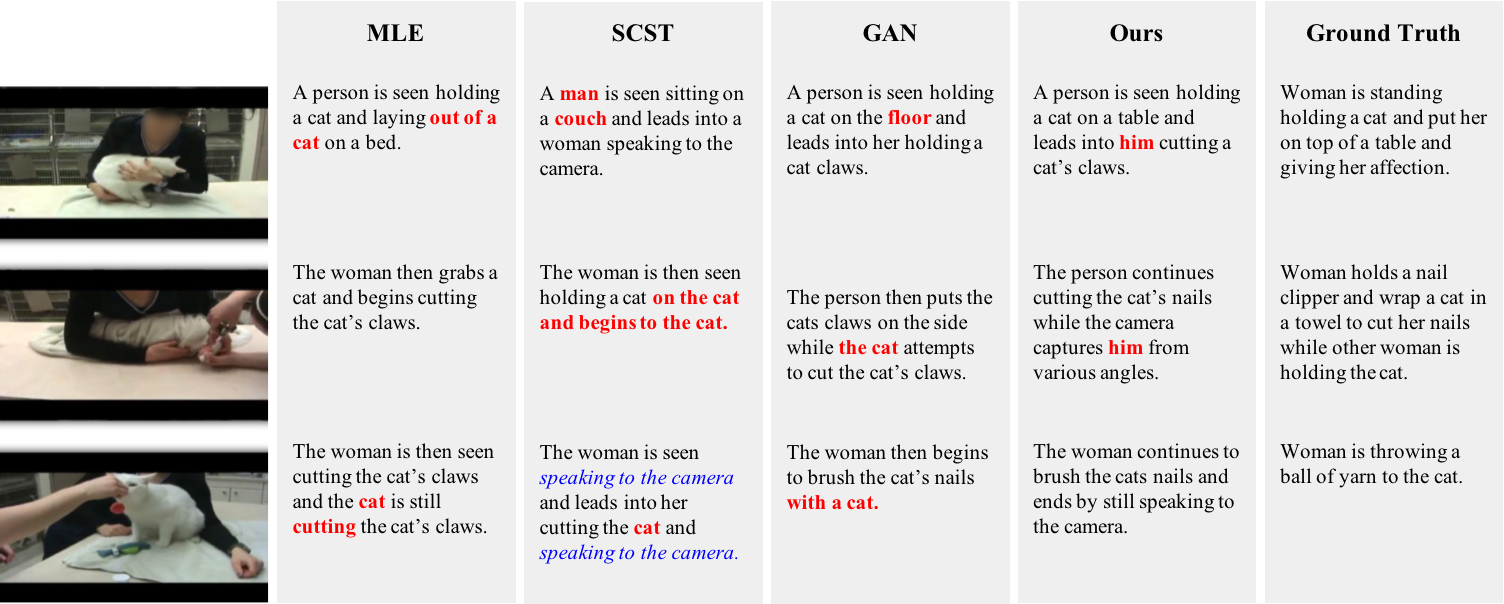} }}%
\caption{Comparison of our approach to MLE baseline, SCST, and GAN. Red/bold indicates content errors, blue/italic indicates repetitive patterns.}
\label{fig:gan_supp}
\end{center}
\end{figure*}

We provide additional examples comparing our approach to SCST and GAN in Figure~\ref{fig:gan_supp}, further illustrating how adversarial inference improves over adversarial training in terms of correctness and fluency. Again, our approach leads to mentioning important concepts, such as \eg ``tai chi''. SCST results in ungrammatical sentence endings (\eg ``a game of'', ``begins to the camera'').

\begin{figure*}%
\scriptsize
\begin{center}
\subfloat[]{{\includegraphics[width=12cm]{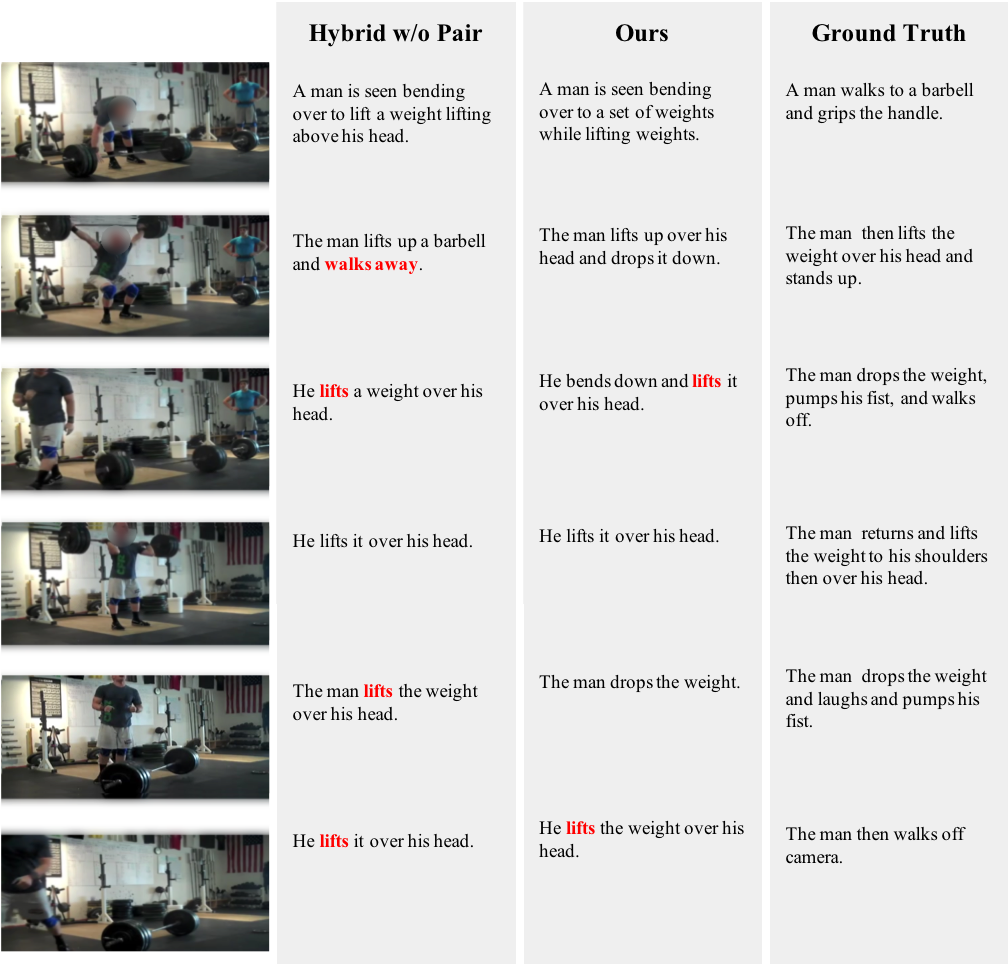} }}%
\quad
    \subfloat[]{{\includegraphics[width=12cm]{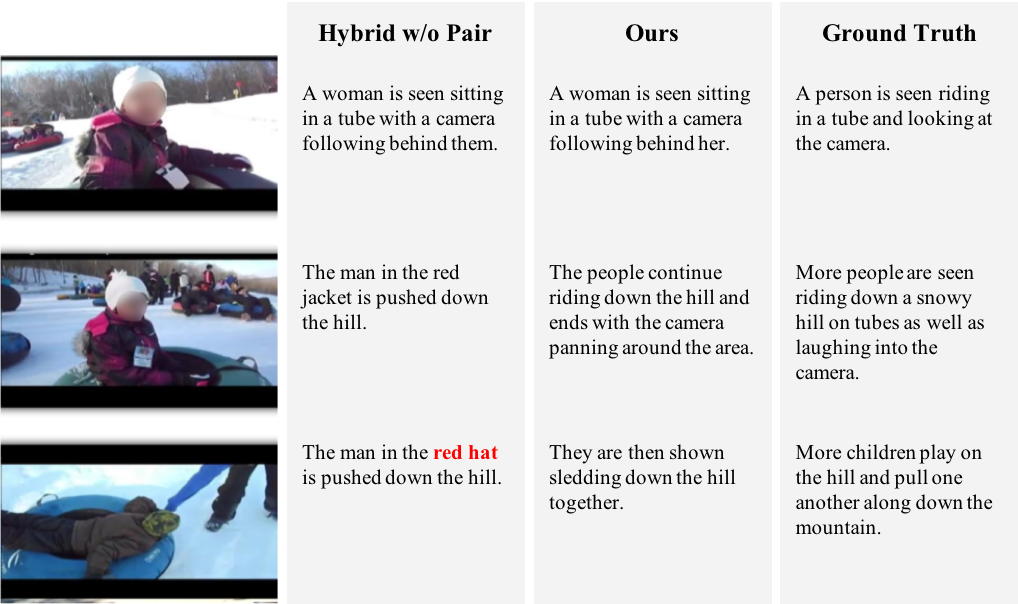} }}%
\caption{Effect of Pairwise Discriminator term in our approach. Red/bold indicates content errors. While both models in a) are not perfectly aligned with ground truth descriptions, the one without pairwise discriminator keeps repeating \textit{lifts a weight} and fails to mention that the man \textit{drops the weight}. Similarly in b), the model without pairwise discriminator mentions that man is \textit{pushed down the hill} twice in a row, while ours avoids generating similar descriptions but more diverse phrases within the paragraph such as \textit{continue riding down the hill} and \textit{shown sledding down the hill together} .}
\label{fig:pair_supp}
\end{center}
\end{figure*}

We also show the effect of our Pairwise Discriminator in Figure~\ref{fig:pair_supp}. As we see, an additional consistency score between sentences helps us obtain less redundant and sometimes more correct predictions (\eg in (a) the hybrid w/o pair never mentions dropping the weights).

\begin{figure*}[t]
\scriptsize
\begin{center}
    \subfloat[]{{\includegraphics[width=13cm]{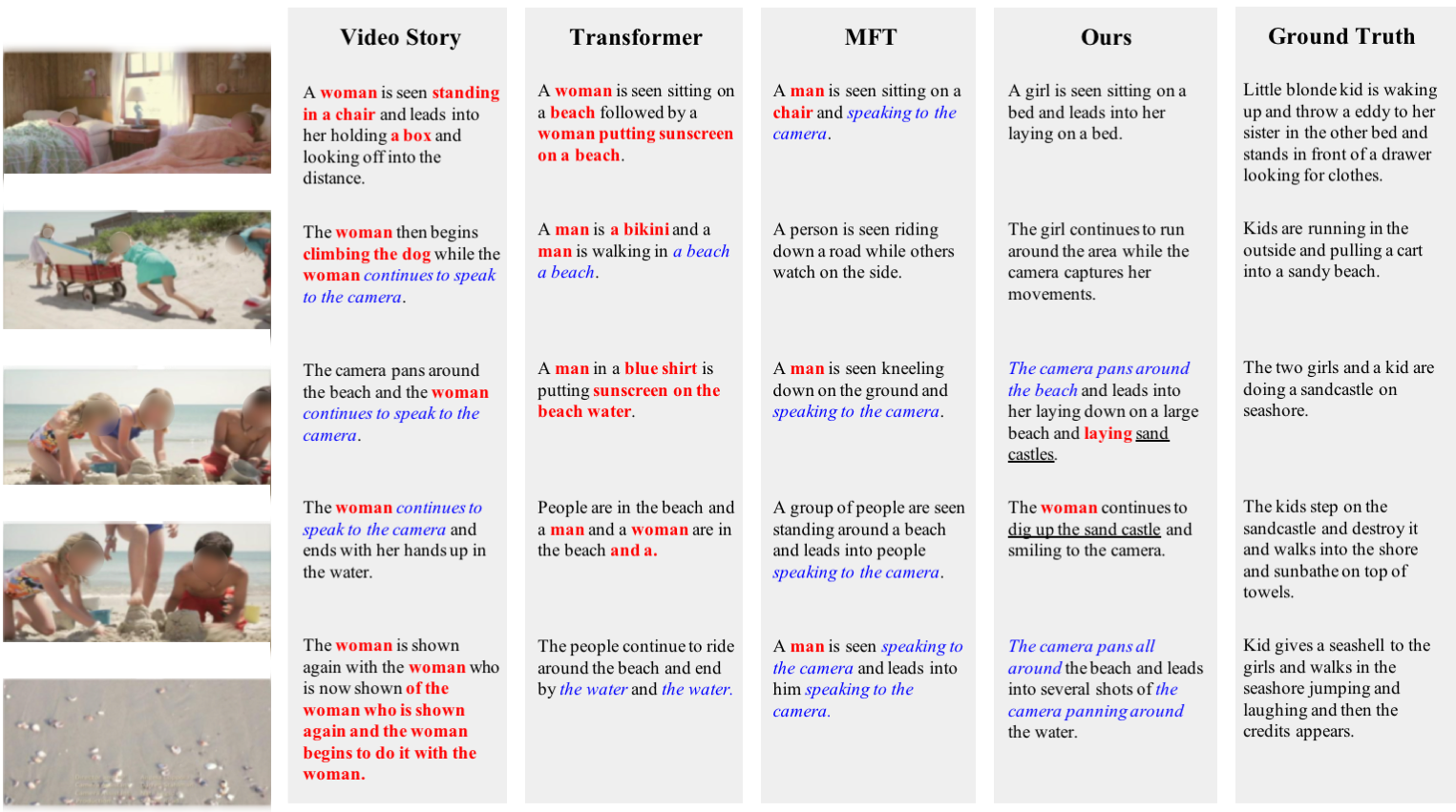} }}%
\quad
    \subfloat[]{{\includegraphics[width=13cm]{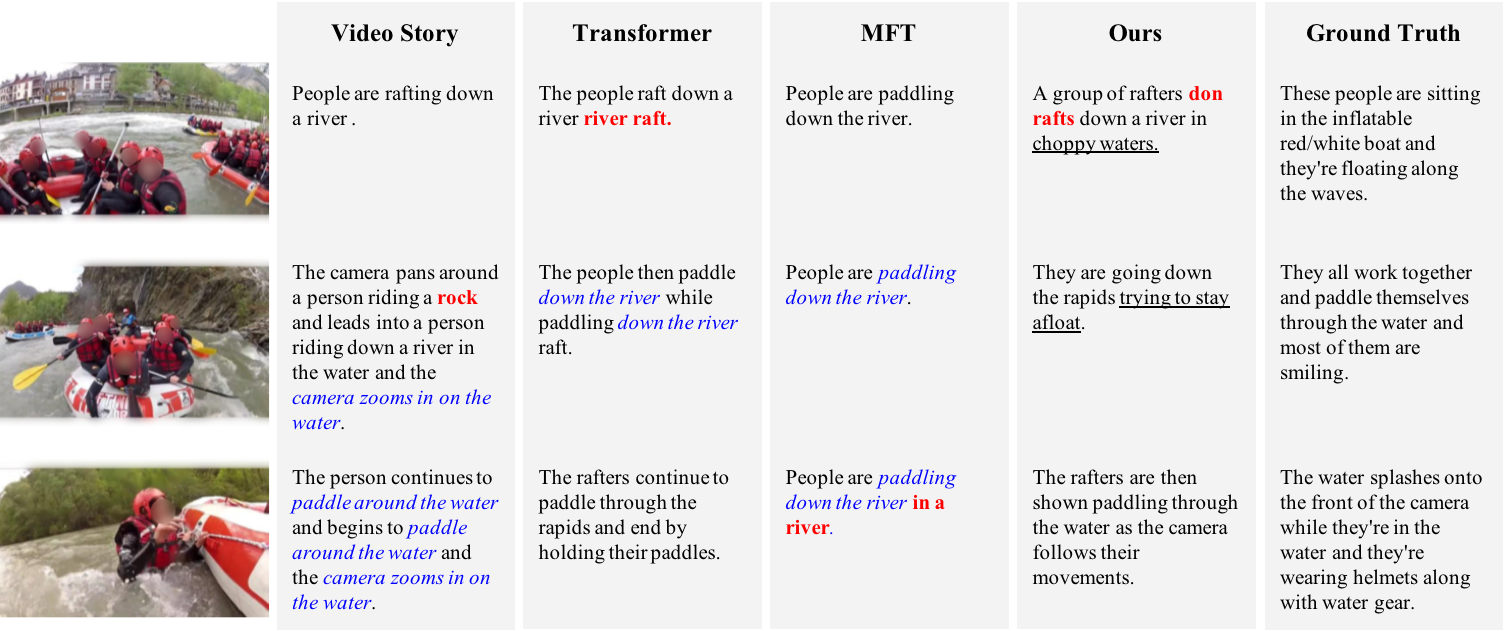} }}%
\quad
    \subfloat[]{{\includegraphics[width=13cm]{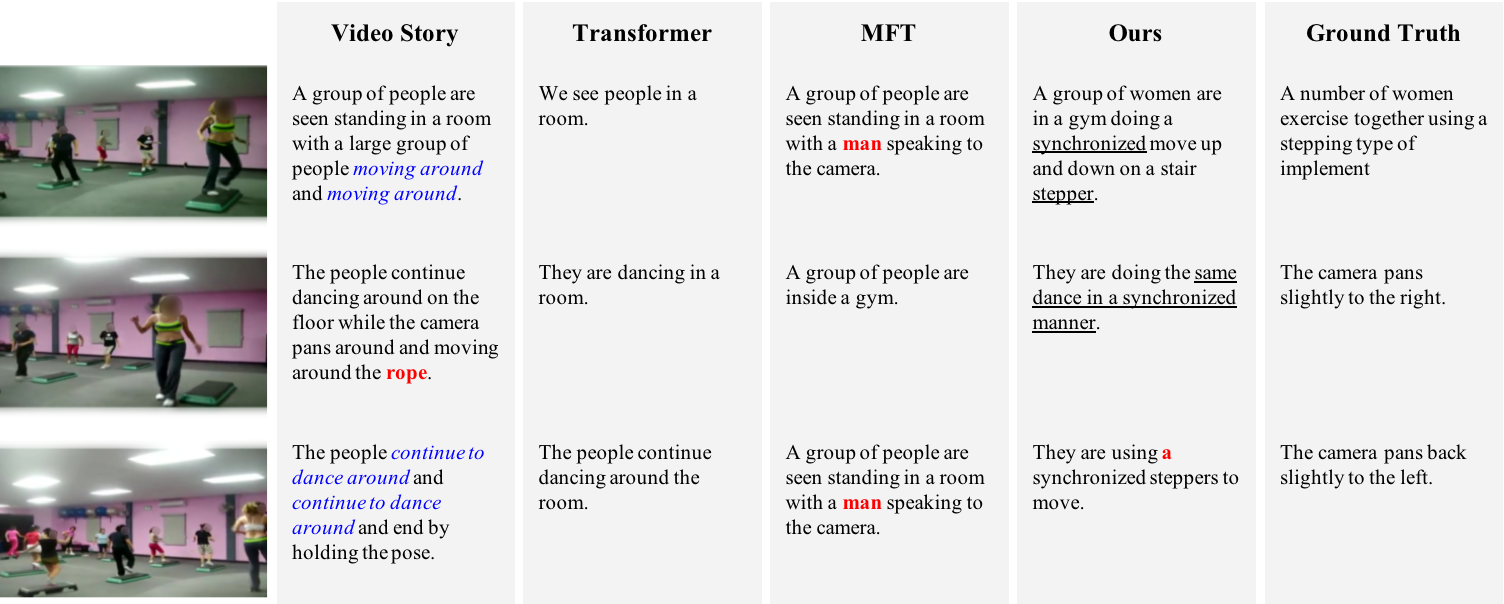} }}%
\caption{Comparison of our approach to state-of-the-art video description approaches (VideoStory  \cite{gella2018dataset}, Transformer \cite{zhou2018end}, MoveForwardTell \cite{xiong2018move}). Red/bold indicates content errors, blue/italic indicates repetitive patterns.}
\label{fig:sota_supp}
\end{center}
\end{figure*}

\subsection{Comparison to State-of-the-Art}
Figure~\ref{fig:sota_supp} provides a comparison of descriptions obtained by our approach to three recent video description models (VideoStory \cite{gella2018dataset}, Transformer \cite{zhou2018end}, MoveForwardTell \cite{xiong2018move}). While the state-of-the-art models are often able to capture the relevant visual information, they are still prone to issues like repetition, lack of diverse and precise content as well as content errors. In particular, VideoStory and MoveForwardTell suffer from the dominant language prior and repeatedly mention ``the camera'', making the stories less informative and specific to the events in the video. Despite having less repeating contents and high scores in language metrics, the Transformer model is prone to produce incoherent phrases \eg ``a man is a bikini'' or ``putting sunscreen on the beach water'', and ungrammatical endings, \eg ``and a'' in (a). On the other hand, our model captures the visual content more precisely, \eg in the top example it refers to the subject as a ``girl'', pointing out that the girl is ``laying on a bed'', correctly recognizing ``sand castles'', etc. Besides, unlike prior work, our approach mentions important video relevant concepts (\eg  ``choppy waters'', ``rapids'', ``afloat'' in (b); ``synchronized'', ``stepper'' in (c)). Overall, we see more diversity and less repetitiveness, along with more accurate description of video content. We note that there is still a large room for improvement w.r.t. the human ground-truth descriptions.

\begin{figure*}[t]
\scriptsize
\begin{center}
    \subfloat[]{{\includegraphics[width=13cm]{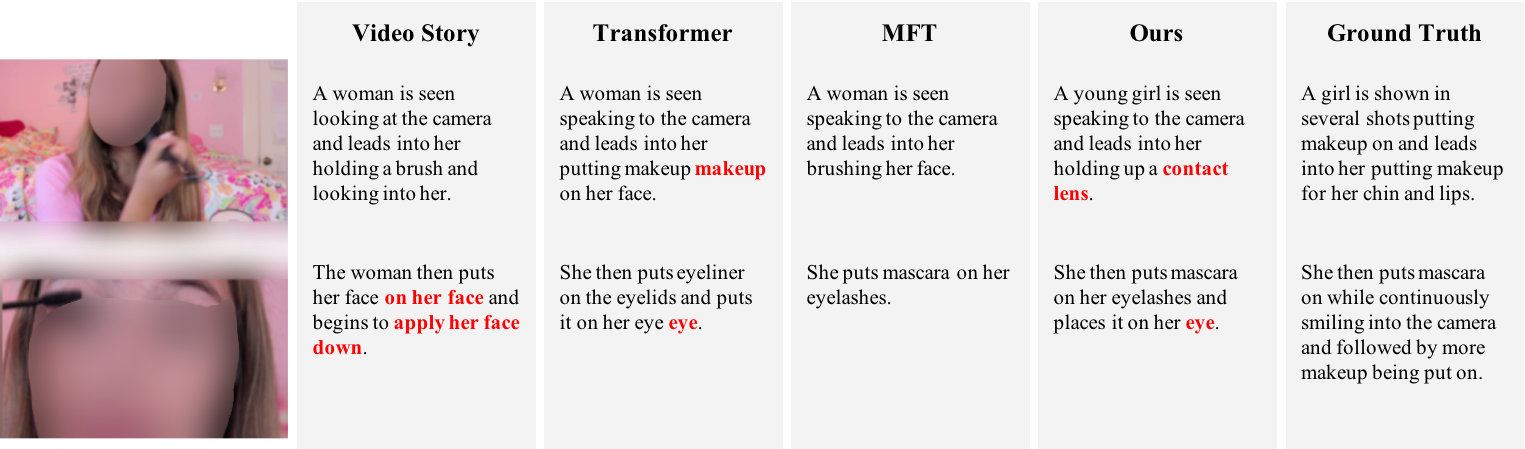} }}%
\quad
    \subfloat[]{{\includegraphics[width=13cm]{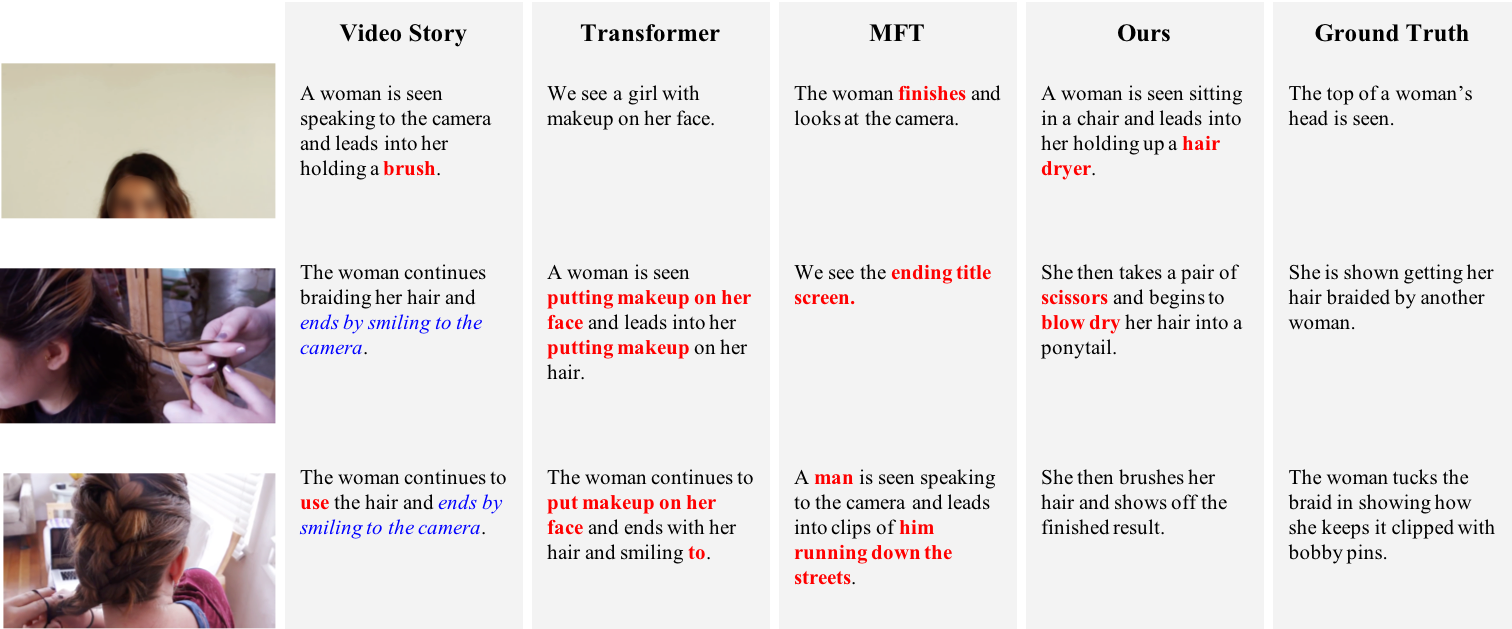} }}%
\quad
    \subfloat[]{{\includegraphics[width=13cm]{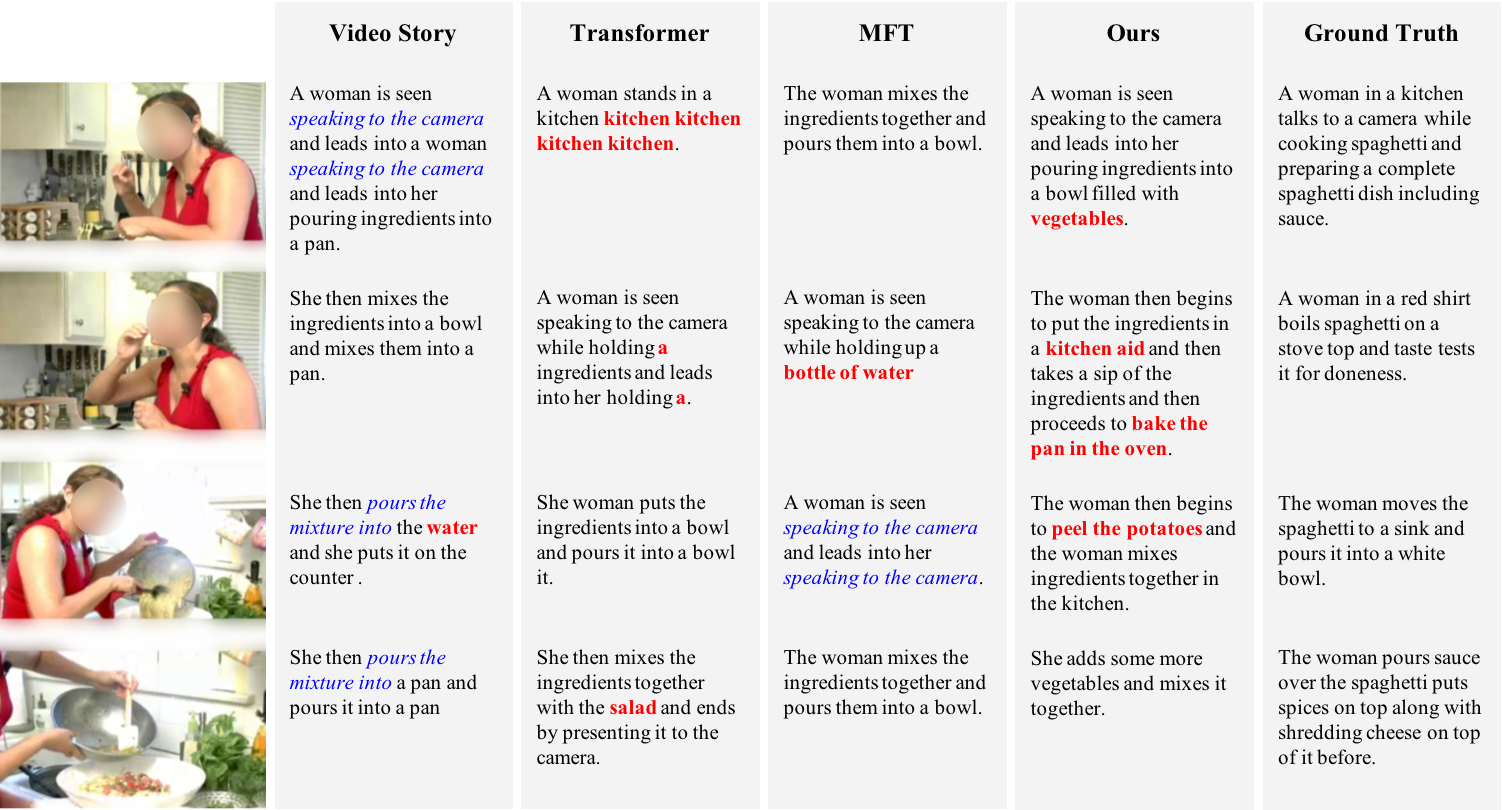} }}%
\caption{Failure cases of our approach and state-of-the-art video description approaches (VideoStory  \cite{gella2018dataset}, Transformer \cite{zhou2018end}, MoveForwardTell \cite{xiong2018move}). Red/bold indicates content errors, blue/italic indicates repetitive patterns.}
\label{fig:failure}
\end{center}
\end{figure*}

\subsection{Failure Analysis}
Finally, we analyze failures of our approach. As shown in the previous examples, our model is not free of errors, \eg it hallucinates an ice cream ``cone'' (Figure~\ref{fig:examples_ours} (a)), incorrectly mentions ``showing off her new york'' (Figure~\ref{fig:examples_ours} (c)), predicts ``man'' instead of a woman (Figure~\ref{fig:gan_supp} (b)) and ``woman'' instead of a child (Figure~\ref{fig:sota_supp} (a)) or ``lifting'' instead of ``dropping'' (Figure~\ref{fig:pair_supp} (a)), etc. It is also still prone to some repetition (\eg Figure~\ref{fig:gan_supp} (a), (b), Figure~\ref{fig:sota_supp} (a)). 
Overall, however, our captions improve over those of the baselines, as supported by our human evaluation.

We include a few additional failure cases in Figure~\ref{fig:failure}, showcasing difficult examples from the ActivityNet Captions dataset. In particular, fine-grained activities that involve small objects are hard, \eg our model confuses applying makeup with inserting a contact lens in Figure~\ref{fig:failure} (a), incorrectly mentions a ``hair dryer'' and ``scissors'' in Figure~\ref{fig:failure} (b), and ``vegetables'' and ``potatos'' in Figure~\ref{fig:failure} (c). The other methods are also struggling on these challenging videos, by either making errors or lacking detail, showing that there is still a long way to go towards solving multi-sentence video description in the wild.

\clearpage

{\small
\bibliographystyle{ieee_fullname}
\bibliography{biblioLong,egbib}
}

\end{document}